\documentclass{article} 
\usepackage{iclr2025_conference,times}
\usepackage{scalefnt,letltxmacro}
\LetLtxMacro{\oldtextsc}{\textsc}
\renewcommand{\textsc}[1]{\oldtextsc{\scalefont{1.10}#1}}
\usepackage[acronym,nowarn]{glossaries}

\glsdisablehyper
\makeglossaries
\usepackage{xspace}
\usepackage{float}
\usepackage{physics}
\usepackage[normalem]{ulem}

\usepackage{enumitem}
\usepackage{amssymb}
\usepackage{mathtools}
\usepackage{amsfonts}
\usepackage{amsmath}
\usepackage{amsthm,thmtools}
\usepackage{booktabs}
\usepackage[]{microtype}

\usepackage[linesnumbered,ruled,vlined]{algorithm2e}

\SetCommentSty{mycommfont}
\SetKwInput{KwInput}{Input}
\SetKwInput{KwOutput}{Output}

\usepackage{hyperref}
\usepackage{svg}

\usepackage[capitalize,nameinlink,]{cleveref} 
\crefname{section}{\S}{\S\S}
\Crefname{section}{\S}{\S\S}

\makeatletter
\@ifundefined{c@rownum}{%
	\let\c@rownum\rownum
}{}
\@ifundefined{therownum}{%
	\def\therownum{\@arabic\rownum}%
}{}
\makeatother

\makeatletter
\newcommand*{\addFileDependency}[1]{
	\typeout{(#1)}
	\@addtofilelist{#1}
	\IfFileExists{#1}{}{\typeout{No file #1.}}
}
\makeatother

\usepackage[font=footnotesize,labelfont=bf,tableposition=top]{caption}
\usepackage{tikz}
\usetikzlibrary{fit}
\usepackage{graphicx}
\usepackage[]{subcaption}   

\usepackage{pgfplots}
\pgfplotsset{compat=1.6}
\usepgfplotslibrary{groupplots}
\tikzstyle{every picture}+=[font=\sffamily]
\tikzstyle{optimized} = [circle,fill=white,draw=black, dashed,inner sep=1pt, minimum size=20pt, font=\fontsize{10}{10}\selectfont, node distance=1]
\pgfkeys{/pgf/number format/.cd,1000 sep={}}
\pgfplotsset{
	tick label style = {font=\sffamily},
	every axis label/.append style={font=\sffamily},
	typeset ticklabels with strut,
}
\pgfplotsset{every axis/.append style={
			every x tick label/.append style={font=\fontsize{6pt}{6pt}\sffamily, yshift=.5ex,},
			every y tick label/.append style={font=\fontsize{6pt}{6pt}\sffamily, xshift=.5ex},
			every y label/.append style={xshift=10ex, font=\sffamily},
			every x label/.append style={yshift=3ex, font=\sffamily},
			every title/.append style={font=\sffamily}
		},
}
\pgfplotsset{
	xticklabel={$\mathsf{\pgfmathprintnumber{\tick}}$},
	yticklabel={$\mathsf{\pgfmathprintnumber{\tick}}$},
}
\pgfplotsset{every axis title/.append style={yshift=-1ex}}
\newlength\figureheight
\newlength\figurewidth
\usepgfplotslibrary{external}
\tikzexternaldisable

\usepackage[colorinlistoftodos,
	textsize=scriptsize,
	linecolor=red!30,
	bordercolor=red!30,
	backgroundcolor=red!10]{todonotes}
\renewcommand{\todo}[2][]{\tikzexternaldisable\@todo[#1]{#2}\tikzexternalenable}

\usepackage[most]{tcolorbox}
{\begin{tcolorbox}[toprule=2mm,left=4pt,right=4pt,top=0pt,bottom=1pt,boxsep=2pt, before skip = 2ex, after skip =2ex]}{\end{tcolorbox}}
\tcbset{boxsep=4pt,left=2pt,right=2pt,top=-0pt,bottom=0pt}

\usepackage{url}

\usepackage{subcaption}

\usepackage{tikz}
\usepackage{graphicx}
\usepackage{xcolor}
\usepackage{amsmath}
\usepackage{amssymb}
\usepackage{bm}
\usepackage{amsthm} 
\usepackage{nicefrac}

\declaretheorem[name=Theorem]{theorem}
\declaretheorem[name=Lemma]{lemma}

\declaretheorem[name=Assumption]{assumption}

\declaretheorem[name=Problem]{problem}

\usepackage{adjustbox}
\usepackage{multirow}
\usepackage{mathtools}

\usepackage{xspace}

\newacronym{FID}{\textsc{fid}}{Fr\'echet Inception Distance}
\newacronym{VAE}{\textsc{vae}}{Variational Autoencoder}
\newacronym{GAN}{\textsc{gan}}{Generative Adversarial Network}
\newacronym{SDE}{\textsc{sde}}{Stochastic Differential Equation}
\newacronym{ELBO}{\textsc{elbo}}{evidence lower bound}
\newacronym{KL}{\textsc{kl}}{Kullback-Leibler}
\newacronym{MLP}{\textsc{mlp}}{multilayer perceptron}
\newacronym{NN}{\textsc{nn}}{neural network}
\newacronym{FDP}{\textsc{fdp}}{Functional Diffusion Process}
\newacronym{NFO}{\textsc{nfo}}{Neural Fourier Operator}
\newacronym{INR}{\textsc{inr}}{implicit neural representation}
\newacronym{LSGM}{\textsc{lsgm}}{Generative Modeling in Latent Space}
\newacronym{IDIFF}{\textsc{$\infty$-diff}}{Infinite Diffusion}
\newacronym{FD2F}{\textsc{fd2f}}{From Data To Functa}
\newacronym{SBD}{\textsc{sbd}}{Score Based Diffusion}
\newacronym{EDM}{\textsc{edm-g++}}{Generative Process with Discriminator Guidance}
\newacronym{NLF}{\textsc{nlf}}{Nonlinear Filtering}

\newcommand{\g}{\,|\,}

\DeclarePairedDelimiterX{\infdivx}[2]{[}{]}{%
#1\;\delimsize\|\;#2%
}

\newcommand{\cB}{\mathcal{B}}
\newcommand{\cC}{\mathcal{C}}
\newcommand{\cS}{\mathcal{S}}

\newcommand{\cP}{\mathcal{P}}

\newcommand{\cR}{\mathcal{R}}
\newcommand{\cF}{\mathcal{F}}
\newcommand{\cI}{\mathcal{I}}

\newcommand{\cG}{\mathcal{G}}

\newcommand{\E}{\mathbb{E}}

\newcommand{\defeq}{\stackrel{\text{\tiny def}}{=}}

\newcommand{\piphi}{ { {\pi} } }
\newcommand{\piphit}{ { {\pi}_t } }
\newcommand{\piphis}{ { {\pi}_s } }

\newcommand{\gX}{\phi}

\newcommand{\rX}{V}

\newcommand{\filt}{\cF}

\newcommand{\ctF}{\cR}

\newcommand{\ccU}{(\Omega,\filt,\PO)}
\newcommand{\cctU}{(\Omega,\ctF_T,\PT)}

\newcommand{\PT}{{\mathrm{Q}^{\ctF}}}
\newcommand{\PQ}{{\mathrm{Q}}}
\newcommand{\PO}{\mathrm{P}}

\newcommand{\mt}{m_t}
\newcommand{\ft}{\frac{\dd \log \mt}{\dd t}}

\newcommand{\scalprod}[2]{\ensuremath{\langle #1, #2\rangle}} %

\definecolor{darkgreen}{rgb}{0.0, 0.5, 0.0}
\usepackage{url}

\title{Latent Abstractions in Generative Diffusion Models}

\author{
Giulio Franzese$^{1}$,
Mattia Martini$^{2}$, Giulio Corallo$^{1,3}$, Paolo Papotti $^{1}$, Pietro Michiardi$^{1}$\\
{$^{1}$ Department of Data Science, EURECOM, Biot, France,}\\{$^{2}$ Université Côte d’Azur, CNRS, Laboratoire J. A. Dieudonné , Nice, France}\\{$^{3}$ SAP Labs, Mougins, France}
}

\iclrfinalcopy 
\begin{document}

\maketitle

\begin{abstract}
In this work we study how diffusion-based generative models produce high-dimensional data, such as an image, by implicitly relying on a manifestation of a low-dimensional set of latent abstractions, that guide the generative process. We present a novel theoretical framework that extends \gls{NLF}, and that offers a unique perspective on \acrshort{SDE}-based generative models. 
The development of our theory relies on a novel formulation of the joint (state and measurement) dynamics, and an information-theoretic measure of the influence of the system state on the measurement process. 
According to our theory, diffusion models can be cast as a system of \acrshort{SDE}, describing a non-linear filter in which the evolution of unobservable latent abstractions steers the dynamics of an observable measurement process (corresponding to the generative pathways). In addition, we present an empirical study to validate our theory and previous empirical results on the emergence of latent abstractions at different stages of the generative process.
\end{abstract}

\section{Introduction}\label{sec:intro}
Generative models have become a cornerstone of modern machine learning, offering powerful methods for synthesizing high-quality data across various domains such as image and video synthesis \citep{dhariwal2021,ho2022,he2022}, natural language processing~\citep{li2022diffusionlm,he-etal-2023-diffusionbert, gulrajani2023likelihoodbased,loudiscrete}, audio generation \citep{kong2021,liu2022}, and molecular structures and general 3D shapes \citep{trippe2022, hoogeboom22, luo2021, zeng2022}, to name a few. These models transform an initial distribution, which is simple to sample from, into one that approximates the data distribution. Among these, diffusion-based models designed through the lenses of \glspl{SDE}  \citep{song2021a,ho2020,albergo2023stochastic} have gained popularity due to their ability to generate realistic and diverse data samples through a series of stochastic transformations. 

In such models, the data generation process, as described by a substantial body of empirical research \citep{chen2023surfacestatisticsscenerepresentations,linhardt2024analysis,tang2023emergent}, appears to develop according to distinct stages: high-level semantics emerge first, followed by the incorporation of low-level details, culminating in a refinement (denoising) phase. Despite ample evidence, a comprehensive theoretical framework for modeling these dynamics remains underexplored.
Indeed, despite recent work on \gls{SDE}-based generative models \citep{berner2022optimal,richter2023improved,ye2022first,raginsky2024variational} shed new lights on such models, they fall short of explicitly investigating the emergence of abstract representations in the generative process. We address this gap by establishing a new framework for elucidating how generative models construct and leverage latent abstractions, approached through the paradigm of \gls{NLF} \citep{bain2009fundamentals,van2007filtering,kutschireiter2020hitchhiker}.

\gls{NLF} is used across diverse engineering domains \citep{bain2009fundamentals}, as it provides robust methodologies for the estimation and prediction of a system's state amidst uncertainty and noise.
\gls{NLF} enables the inference of dynamic latent variables that define the system state based on observed data, offering a Bayesian interpretation of state evolution and the ability to incorporate stochastic system dynamics. 
The problem we consider is the following: an \textit{unobservable} random variable $X$ is measured through a noisy continuous-time process $Y_t$, wherein the influence of $X$ on the noisy process is described by an observation function $H$, with the noise component modeled as a Brownian motion term. The goal is to estimate the a-posteriori measure $\pi_t$ of the variable $X$ given the entire historical trajectory of the measurement process $Y_t$. 


In this work, we establish a connection between \gls{SDE}-based generative models and \gls{NLF} by observing that they can be interpreted as \textit{simulations} of \gls{NLF} dynamics.
In our framework, the latent abstraction, which corresponds to certain real-world properties within the scope of classical nonlinear filtering and remains unaffected in a \textit{causal} manner by the posterior process $\pi_t$, is implicitly simulated and iteratively refined. We explore the connection between latent abstractions and the a-posteriori process, through the concept of \textit{filtrations} -- broadly defined as collections of progressively increasing information sets -- and offer a rigorous theory to study the emergence and influence of latent abstractions throughout the data generation process.
Our theoretical contributions unfold as follows.

In \Cref{sec:NLF} we show how to reformulate classical \gls{NLF} results such that the measurement process is the only available information, and derive the corresponding dynamics of both the latent abstraction and the measurement process. These results are summarized in \Cref{innovation_theo} and \Cref{theo_girsa}.

Given the new dynamics, in \Cref{kushner_eq} we show how to estimate the a-posteriori measure of the \gls{NLF} model, and present a novel derivation to compute the mutual information between the measurement process and random variables derived from a transformation of the latent abstractions in \Cref{theo_mi}. Finally, we show in \Cref{sufficient_statistic}, that the a-posteriori measure is a sufficient statistics for any random variable derived from the latent abstractions, when only having access to the measurement process.

Building on these general results, in \Cref{sec:GEN} we present a novel perspective on continuous-time score-based diffusion models, which is summarised in \Cref{eq:system}. We propose to view such generative models as \gls{NLF} simulators that progress in two stages: first our model updates the a-posteriori measure representing a sufficient statistics of the latent abstractions, second, it uses a projection of the a-posteriori measure to update the measurement process. Such intuitive understanding is the result of several fundamental steps. 
In \Cref{anders_thm} and \Cref{theo:connection}, we show that the common view of score-based diffusion models by which they evolve according to forward (noising) and backward (generative) dynamics is compatible with the \gls{NLF} formulation, in which there is no need to distinguish between such phases. In other words, the \gls{NLF} perspective of \Cref{eq:system} is a valid generative model. 
In \Cref{sec:lineardiff}, we provide additional results (see \Cref{anders_thm2}), focusing on the specific case of linear diffusion models, which are the most popular instance of score-based generative models in use today.
In \Cref{sec:summary}, we summarize the main intuitions behind our \gls{NLF} framework.

Our results explain, by means of a theoretically sound framework, the emergence of latent abstractions that has been observed by a large body of empirical work \citep{bisk2020experience,bender2020climbing,li2022emergent,park2023understanding,kwon2023diffusionmodelssemanticlatent,chen2023surfacestatisticsscenerepresentations,linhardt2024analysis,tang2023emergent,xiang2023denoising,haas2024discoveringinterpretabledirectionssemantic}. The closest research to our findings is discussed in \citep{sclocchi2024phasetransitiondiffusionmodels}, albeit from a different mathematical perspective. 
To root our theoretical results in additional empirical evidence, we conclude our work in \Cref{sec:empirical} with a series of experiments on score-based generative models~\citep{song2021a}, where we 1) validate existing probing techniques to measure the emergence of latent abstractions, 2) compute the mutual information as derived in our framework, and show that it is a suitable approach to measure the relation between the generative process and latent abstractions, 3) introduce a new measurement protocol to further confirm the connections between our theory, and how practical diffusion-based generative models operate.



\section{Nonlinear Filtering}\label{sec:NLF}

Consider two random variables $Y_t$ and $X$, corresponding to a stochastic\textbf{ measurement} process ($Y_t$) of some underlying \textbf{latent abstraction} ($X$).  We construct our universe sample space $\Omega$ as the combination of the space of continuous functions in the interval $[0,T]$ ($T\in\mathbb{R}^+$) and of a complete separable metric space $\cS$, i.e., $\Omega=\cC([0, T],\mathbb{R}^N)\times \cS$. On this space, we consider the joint \textit{canonical} process $Z_t(\omega)=[Y_t,X]=[\omega_t^y,\omega^x]$ for all $\omega \in \Omega$, with $\omega=[\omega^y,\omega^x]$. In this work we indicate with $\sigma(\cdot)$ sigma-algebras. Consider the growing filtration naturally induced by the canonical process $\cF^{Y,X}_t=\sigma( Y_{0\leq s\leq t},X )$ (a short-hand for $\sigma(\sigma(Y_{0\leq s\leq t})\cup\sigma(X))$), and define $\cF=\cF^{Y,X}_{T}$.
We build the probability triplet $\ccU$, where the probability measure $\PO$ is selected such that the process $\{Z_{0\leq t\leq T},\cF^{Y,X}_{0\leq t\leq T}\}$ has the following \gls{SDE} representation
\begin{equation}\label{eq:param_sde}
    Y_t=Y_0+\int_0^t H(Y_s,X,s) \dd s+ W_t,
\end{equation}
where $\{W_{0\leq t\leq T},\cF^{Y,X}_{0\leq t\leq T}\}$ is a Brownian motion with initial value 0
and $H: \Omega \times [0,T] \rightarrow \mathbb{R}^N$
is an \textit{observation} process. 
All standard technical assumptions are available in \Cref{app:assumptions}.


Next, we provide the necessary background on \gls{NLF}, to pave the way for understanding its connection with the generative models of interest.
The most important building block of the \gls{NLF} literature is represented by the \textbf{conditional probability measure} $\PO[ X\in A\g \cF^{Y}_t]$ (notice the reduced filtration $\cF^Y_t\subset \cF^{Y,X}_t$), which summarizes, a-posteriori, the distribution of $X$ given observations of the measurement process until time $t$, that is, $Y_{0\leq s\leq t}$. 


\begin{theorem}\label{base}[Thm 2.1 \citep{bain2009fundamentals}]
    Consider the probability triplet $\ccU$, the metric space $\cS$ and its Borel sigma-algebra $\cB(\cS)$.
    There exists a (probability measure valued $\cP(\cS)$) process $\{\piphi_{0\leq t\leq T},\cF^{Y}_{0\leq t\leq T}\}$, with a progressively measurable modification, such that for all $A\in \cB(\cS)$, the conditional probability measure $\PO[ X\in A\g \cF^{Y}_t]$ is well defined and is equal to $\piphit(A)$. 
\end{theorem}

The conditional probability measure is extremely important, as the fundamental goal of nonlinear filtering is the solution of the following problem. Here, we introduce the quantity $\gX$, which is a random variable derived from the latent abstractions $X$. 

\begin{problem}\label{nnfilt}
For any fixed $\gX:\cS \to \mathbb{R}$ bounded and measurable, given knowledge of the measurement process $Y_{0\leq s\leq t}$, compute $ \E_{\PO}[\gX(X)\g \cF^{Y}_t]$. This amounts to computing 
\begin{equation}\label{bayes_avg}
\langle\piphit,\gX\rangle=\int_{ \cS}\gX(x)\dd\piphit(x).
\end{equation}
\end{problem}




In simple terms, \Cref{nnfilt} involves studying the existence of the a-posteriori measure and the implementation of efficient algorithms for its update, using the flowing stream of incoming information $Y_t$. We first focus our attention on the existence of an analytic expression for the value of the a-posteriori expected measure $\piphit$. 
Then, we quantify the interaction dynamics between observable measurements and $\gX$, through the lenses of mutual information $\cI(Y_{0\leq s\leq t};\gX)$, which is an extension of the problems considered in \citep{newton2008interactive,duncan1970calculation,DUNCAN1971265,mitter2003variational}.

\subsection{Technical preliminaries}

We set the stage of our work by revisiting the measurement process $Y_t$, and express it in a way that does not require access to unobservable information.
Indeed, while $Y_t$ is naturally adapted w.r.t. its own filtration $\cF^Y_t$, and consequently to any other growing filtration $\ctF_t$ such $\cF^{Y,X}_t\supseteq \ctF_t\supseteq \cF^{Y}_t$, the representation in \Cref{eq:param_sde} is in general not adapted, letting aside degenerate cases. 

Let's consider the family of growing filtrations $\ctF_t=\sigma(\ctF_0\cup \sigma(Y_{0\leq s\leq t}-Y_0))$, where $\sigma(Y_0)\subseteq \ctF_0\subseteq \sigma(X,Y_0)$. Intuitively $\ctF_0$ allows to modulate between the two extreme cases of knowing only the initial conditions of the \gls{SDE}, that is $Y_0$, to the case of complete knowledge of the whole latent abstraction $X$, and anything in between. As shown hereafter, the original process $Y_t$ associated to the space $(\Omega, \filt,\PO)$ which solves \Cref{eq:param_sde}, also solves \Cref{genh_sde}, that is adapted on the reduced filtration $\ctF_t$. This allows us to reason about the partial observation of the latent abstraction ($\ctF_0$ vs $\sigma(X,Y_0)$), without incurring in the problem of the measurement process $Y_t$ being statistically dependent of the whole latent abstraction $X$.

Armed with such representation, we study under which change of measure the process $Y_t-Y_0$ behaves as a Brownian motion (\Cref{theo_girsa}). This serves the purpose of simplifying the calculation of the expected value of $\phi$ given $Y_t$, as described in \Cref{nnfilt}. Indeed, if $Y_t-Y_0$ is a Brownian motion independent of $\phi$, its knowledge does not influence our best guess for $\phi$, i.e. the conditional expected value. Moreover, our alternative representation is instrumental for the efficient and simple computation of the mutual information $\cI(Y_{0\leq s\leq t};\gX)$, where the different measures involved in the Radon-Nikodym derivatives will be compared against the same reference Brownian measures.

The first step to define our representation is provided by the following
\begin{theorem}\label{innovation_theo}\hyperref[proof_innovation_theo]{[Proof].}
Consider the the probability triplet $\ccU$, the process in \Cref{eq:param_sde} defined on it, and the growing filtration $\ctF_t=\sigma(\ctF_0\cup \sigma(Y_{0\leq s\leq t}-Y_0))$. Define a new stochastic process
    \begin{equation}\label{browphi}
        {W}^{\cR}_t\defeq  Y_t-Y_0-\int_0^t \E_{\PO}(H(Y_s,X,s)\g \ctF_s)\dd s.
    \end{equation}
Then, $\{{W}^{\cR}_{0\leq t\leq T},\ctF_{0\leq t\leq T}\}$ is a Brownian motion. 
Notice that if $\ctF_t=\cF^{Y,X}_t$, then ${W}^{\cR}_t=W_t$.
\end{theorem}

Following \cref{innovation_theo}, the process $\{Y_{0\leq t\leq T},\ctF_{0\leq t\leq T}\}$ has \gls{SDE} representation 
\begin{equation}\label{genh_sde}
Y_t=Y_0+\int_0^t \E_{\PO}(H(Y_s,X,s)\g \ctF_s)\dd s + {W}^{\cR}_t.    
\end{equation}



Next, we derive the change of measure necessary for the process $\tilde{W}_t\defeq Y_t-Y_0$ to be a Brownian motion w.r.t to the filtration $\ctF_t$. To do this, we apply the Girsanov theorem \citep{oksendal2003stochastic} to $\tilde{W}_t$ which, in general, admits a $\cR$ -- adapted representation $\int_0^t \E_{\PO}(H(Y_s,X,s)\g \ctF_s)\dd s+{W}^{\cR}_t$.

\begin{theorem} \label{theo_girsa}\hyperref[proof_theo_girsa]{[Proof].}
Define the new probability space $\cctU$ via the measure $\PT(A)=\E_{\PO}\left[ \mathbf{1}(A)({\psi^{\ctF}_T})^{-1}\right]$, for $A\in \ctF_T$, where 
\begin{equation}\label{radonniko}
\psi^{\ctF}_t\defeq \exp(\int_0^t \E_{\PO}[H(Y_s,X,s)\g\ctF_s]\dd Y_s-\frac{1}{2}\int_0^t\norm{\E_{\PO}[H(Y_s,X,s)\g\ctF_s]}^2 \dd s ),
\end{equation}
and  
$$\PT\g_{\ctF_t}=\E_{\PO}\left[ \mathbf{1}(A)\E_\PO[({\psi^{\ctF}_T})^{-1}\g\ctF_t]\right]=\E_{\PO}\left[ \mathbf{1}(A)({\psi^{\ctF}_t})^{-1}\right].$$

Then, the stochastic process $\{\tilde{W}_{0\leq t\leq T}, \ctF_{0\leq t\leq T}\}$ is a Brownian motion on the space $\cctU$.


\end{theorem}



A direct consequence of \cref{theo_girsa} is that the process $\tilde{W}_t$ is independent of any $\ctF_0$ measurable random variable under the measure $\PT$. 
Moreover, it holds that for all $\mathcal{R}'_t\subseteq \mathcal{R}_t$, 
$\PT\g_{\mathcal{R}'_t}=\mathrm{Q}^{\mathcal{R}'}\g_{\mathcal{R}'_t}
$.

\subsection{A-Posteriori Measure and Mutual Information}
As we did in \Cref{sec:NLF} for the process $\pi_t$, here we introduce a new process $\pi^\ctF_t$ which represents the conditional law of $X$ given the filtration $\ctF_t=\sigma(\ctF_0\cup \sigma(Y_{0\leq s\leq t}-Y_0))$. More precisely, for all $A\in \cB(\cS)$, the conditional probability measure $\PO[ X\in A\g \mathcal{R}_t]$ is well defined and is equal to $\pi^\ctF_t(A)$. Moreover, for any $\gX:\cS \to \mathbb{R}$ bounded and measurable, $\E_{\PO}[\gX(X)\g \mathcal{R}_t] = \langle\pi^\ctF_t,\gX\rangle$. Notice that if $\ctF = \cF^Y$ then $\pi^\ctF$ reduces to $\pi$.

Armed with \Cref{theo_girsa}, we are ready to derive the expression for the a-posteriori measure $\pi^\ctF_t$ and the mutual information between observable measurements and the unavailable information about the latent abstractions, that materialize in the random variable $\gX$. 

\begin{theorem}\label{kushner_eq}\hyperref[proof_kushner_eq]{[Proof].} 
The measure-valued process $\pi^\ctF_t$ solves in weak sense (see \Cref{proof_kushner_eq} for a precise definition), the following \gls{SDE}
    \begin{flalign}
        &\pi^\ctF_t=\pi^\ctF_0+\int_0^t \pi^\ctF_s\left(H(Y_s,\cdot,s)-\langle \pi^\ctF_s,H(Y_s,\cdot,s)\rangle\right) \left(\dd Y_s-\langle \pi^\ctF_s,H(Y_s,\cdot,s)\rangle\dd s\right),\label{kush}
    \end{flalign}
where the initial condition $\pi_0$ satisfies $\pi^\ctF_0(A)=\PO[X\in A\g \ctF_0]$ for all $A\in \cB(\cS)$.
\end{theorem}

When $\ctF = \cF^Y$, \Cref{kush} is the well-know Kushner-Stratonovitch (or Fujisaki-Kallianpur-Kunita) equation (see e.g.~\cite{bain2009fundamentals}).
A proof for uniqueness of the solution of \Cref{kush} can be approached by considering the strategies in \citep{fotsa2017nonlinear}, but is outside the scope of this work. The (recursive) expression in \Cref{kush} is particularly useful for engineering purposes since, in general, it is usually not known in which variables $\gX(X)$, representing latent abstractions, we could be interested in. Keeping track of the \textit{whole distribution} $\pi^\ctF_t$ at time $t$ is the most cost-effective solution, as we will show later. 

Our next goal is to quantify the interaction dynamics between observable measurements and latent abstractions that materialize through the variable $\gX(X)$ (from now on we write only $\phi$ for the sake of brevity): in \Cref{theo_mi} we derive the mutual information $\cI(Y_{0\leq s\leq t};\gX)$. 
\begin{theorem}\label{theo_mi}\hyperref[proof_theo_mi]{[Proof]} 
The mutual information between observable measurements $Y_{0\leq s\leq t}$ and $\gX$ is defined as:
    \begin{equation}\label{eq:infodef}
       \cI(Y_{0\leq s\leq t};\gX)\defeq \int\log\frac{\dd \PO_{\#Y_{0\leq s\leq t},\gX} }{\dd \PO_{\#Y_{0\leq s\leq t}}\dd \PO_{\#\gX}} \dd \PO_{\#Y_{0\leq s\leq t},\gX}. 
    \end{equation}
It holds that such quantity is equal to $ \E_{\PO}\left[ \log\frac{\dd \PO\g_{\ctF_t}}{\dd \PO\g_{\cF^{Y}_t}\dd \PO\g_{\sigma(\gX)}} \right]$, which can be simplified as follows:
    \begin{equation}\label{MI_eq}
        \cI(Y_{0};\gX)+ \frac{1}{2}\E_{\PO}\left[ \int_0^t\norm{\E_{\PO}[H(X,Y_s,s)\g \cF^{Y}_s]-\E_{\PO}[H(X,Y_s,s)\g \ctF_s]}^2 \dd s \right].
    \end{equation}


\end{theorem}

The mutual information computed by \Cref{MI_eq} is composed by two elements: first, the mutual information between the initial measurements $Y_0$ and $\gX$, which is typically zero by construction. 
The second term quantifies how much the best prediction of the observation function $H$ is influenced by the extra knowledge of $\gX$, in addition to the measurement history $Y_{0\leq s\leq t}$. 
By adhering to the premise that the conditional expectation of a stochastic variable constitutes the optimal estimator given the conditioning information, the integral on the r.h.s quantifies the expected square difference between predictions, having access to measurements only ($\E_{\PO}[\cdot\g \cF^Y_t]$) and those incorporating additional information ($\E_{\PO}[\cdot\g \cR_t]$).

Even though a precise characterization for general observation functions and and variables $\gX$ is typically out of reach, a \textbf{qualitative} analysis is possible. First, the mutual information between $\gX$ and the measurements depends on \textit{i)} how much the amplitude of $H$ is impacted by knowledge of $\gX$ and \textit{ii)} the \textit{number} of elements of $H$ which are impacted (informally, how much localized vs global is the impact of $\gX$). Second, it is possible to define a hierarchical interpretation about the emergence of the various latent factors: a variable with a local impact can ``\textit{appear}'', in an information theoretic sense, only if the impact of other global variables is resolved, otherwise the remaining uncertainty of the global variables makes knowledge of the local variable irrelevant. In classical diffusion models, this is empirically known \citep{chen2023surfacestatisticsscenerepresentations,linhardt2024analysis,tang2023emergent}, and corresponds to the phenomenon where \textit{semantics emerges before details} (global vs local details in our language).

Now, consider any $\cF^Y_t$ measurable random variable $\tilde{Y}_t$, defined as a mapping to a generic measurable space $(\Psi,\cB(\Psi))$, which means it can also be seen as a process.
The \textit{data processing inequality} states that the mutual information between such $\tilde{Y}$ and $\gX$ will be smaller than the mutual information between the original measurement process and $\gX$. However, it can be shown that all the relevant information about the random variable $\gX$ contained in $\cF^Y_t$ is equivalently contained in the filtering process at time instant $t$, that is $\pi_t$. This is not trivial, since $\pi_t$ is a $\cF^Y_t$-measurable quantity, i.e., $\sigma(\pi_t)\subset \cF^Y_t$. In other words, we show that $\pi_t$ is a \textbf{sufficient statistic} for any $\sigma(X)$ measurable random variable when starting from the measurement process.


\begin{theorem}\hyperref[proof:sufficient_statistic]{[Proof]}\label{sufficient_statistic}
For any $\cF^Y_t$ measurable random variable $\tilde{Y}_t:\Omega\rightarrow \Psi$, the following inequality holds:
\begin{equation}\label{eq:dpi}
    \cI(\tilde{Y};\gX)\leq \cI(Y_{0\leq s\leq t};\gX).
\end{equation}

For a given $t\geq 0$, the measurement process $Y_{0\leq s\leq t}$ and $X$ are \textit{conditionally}-independent given $\pi_t$. This implies that $\PO(A\g \sigma(\pi_t))=\PO(A\g \cF^Y_t),\quad \forall A\in\sigma(X)$. Then
$\cI(Y_{0\leq s\leq t};\gX)=\cI(\pi_t;\gX)$ (i.e. \Cref{eq:dpi} is attained with equality).
\end{theorem}
While $\pi_t$ contains all the relevant information about $\phi$, the same cannot be said about the conditional expectation, i.e. the particular case $\tilde{Y}=\langle\pi_t,\phi\rangle$. Indeed, from \Cref{bayes_avg}, $\langle\pi_t,\phi\rangle$ is obtained as a \textit{transformation} of $\pi_t$ and thus can be interpreted as a $\cF^Y_t$ measurable quantity subject to the constraint of \Cref{eq:dpi}. 
As a particular case, the quantity $\langle \piphit, H\rangle$, of central importance in the construction of generative models \Cref{sec:GEN}, carries in general less information about $\phi$ than the un-projected $\pi_t$.


\section{Generative Modelling}\label{sec:GEN}
We are interested in \textbf{generative models} for a given $\sigma(X)$-measurable random variable $\rX$.
An intuitive illustration of how data generation works according to our framework is as follows. Consider, for example, the image domain, and the availability of a rendering engine that takes as an input a computer program describing a scene (coordinates of objects, textures, light sources, auxiliary labels, etc ...) and that produces an output image of the scene. In a similar vein, a generative model learns how to use latent variables (which are not explicitly provided in input, but rather implicitly learned through training) to generate an image.
For such model to work, one valid strategy is to consider an \gls{SDE} in the form of \Cref{eq:param_sde} where the following holds\footnote{
 From a strict technical point of view, \Cref{assumption1} might be incompatible with other assumptions in \Cref{app:assumptions}, or proving compatibility could require particular effort. Such details are discussed in \Cref{sec:technicalnote}.
 }\label{foot1}.  
\begin{assumption}\label{assumption1}
The stochastic process $Y_t$ satisfies $Y_T=\rX,\quad \PO-a.s.$
\end{assumption} 
Then, we could numerically simulate the dynamics of \Cref{eq:param_sde} until time $T$. Indeed, starting from initial conditions $Y_0$, we could obtain $Y_T$ that, under \Cref{assumption1}, is precisely $\rX$. Unfortunately, such a simple idea requires \textit{explicit access} to $X$, as it is evident from \Cref{eq:param_sde}. In mathematical terms, \Cref{eq:param_sde} is adapted to the filtration $\cF^{Y,X}_t$. However, we have shown how to reduce the available information to account only for historical values of $Y_t$. Then, we can combine the result in \Cref{kushner_eq} with \Cref{innovation_theo} and re-interpret \Cref{genh_sde}, which is a valid generative model, as 
\begin{equation}\label{eq:system}
\begin{cases}
\pi_t=\pi_0+\int_0^t \pi_s\left(H-\langle \piphis,H\rangle\right) \left(\dd Y_s-\langle \piphis,H\rangle\dd s\right),\\
Y_t=Y_0+\int_0^t \langle \piphis,H\rangle \dd s+{W}^{\cF^Y}_t,    
\end{cases}
\end{equation}   
where $H$ denotes $H(Y_s,\cdot,s)$.
Explicit simulation of \Cref{eq:system} only requires knowledge of the whole history of the measurement process: provided \Cref{assumption1} holds, it allows generation of a sample of the random variable $\rX$. 


Although the discussion in this work includes a large class of observation functions, we focus on the particular case of generative diffusion models \citep{song2021a}. Typically, such models are presented through the lenses of a forward noising process and backward (in time) \glspl{SDE}, following the intuition of \citet{anderson1982reverse}. Next, according to the framework we introduce in this work, we reinterpret such models under the perspective of enlargement of filtrations.

Consider the \textit{reversed} process $\hat{Y}_t\defeq Y_{T-t}$ defined on $\ccU$ and the corresponding filtration $\cF_t^{\hat{Y}}\defeq\sigma(\hat{Y}_{0\leq s\leq t})$. The measure $\PO$ is selected such that the process $\hat{Y}_t$ has $\cF_t^{\hat{Y}}$--adapted expression
\begin{equation}\label{eq:diffmod_gen}
\hat{Y}_t=\rX+\int\limits_{0}^t F(\hat{Y}_{s},s)\dd s+\hat{W}_t,
\end{equation}
where $\{\hat{W}_t,\cF_t^{\hat{Y}}\}$ is a Brownian motion. Then, \Cref{assumption1} is valid since $Y_T=\hat{Y}_0=\rX$. 
Note that \Cref{eq:diffmod_gen}, albeit with a different notation, is reminiscent of the forward \gls{SDE} that is typically used as the starting point to illustrate score-based generative models~\citep{song2021a}. In particular, $F(\cdot)$ corresponds to the drift term of such a diffusion \gls{SDE}.

\Cref{eq:diffmod_gen} is equivalent to $Y_{t}=\rX+\int\limits_{t}^{T} F(Y_{s},T-s)\dd s+\hat{W}_{T-t}$, which is an expression for the process $Y_t$, which is adapted to $\cF^{\hat{Y}}$. This constitutes the first step to derive an equivalent backward (generative) process according to the traditional framework of score-based diffusion models. 
Note that such an equivalent representation is not useful for simulation purposes: the goals of the next steps is to transform it such that it is adapted to $\cF^{Y}$. Indeed, using simple algebra, it holds that 

$$Y_{t}=Y_0-\int\limits_{0}^{t} F(Y_{s},T-s)\dd s+\left(-Y_0+ \rX+\int\limits_{0}^{T} F(Y_{s},T-s)\dd s+\hat{W}_{T-t}\right),$$ 

where the last term in the parentheses is equal to $-\hat{W}_T+\hat{W}_{T-t}$. 

Note that $\cF_t^{Y}=\sigma(\hat{Y}_{T-t\leq s\leq T})$. 
Since $\sigma(\hat{Y}_{T-t\leq s\leq T})=\sigma(\hat{W}_{T-t\leq s\leq T})\cup \sigma(\hat{Y}_{T-t})$, we can apply the result in \citep{pardoux2006grossissement} (Thm 2.2) to claim the following: $-\hat{W}_T+\hat{W}_{T-t}-\int_0^t\nabla\log \hat{p}(Y_s,T-s)\dd s$ is a Brownian motion adapted to $\cF^Y_t$, where this time $\PO(\hat{Y}_t\in \dd y)=\hat{p}(y,t)\dd y$. Then \citep{pardoux2006grossissement}
\begin{theorem}\label{anders_thm}
    Consider the stochastic process $Y_t$ which solves \Cref{eq:diffmod_gen}. The same stochastic process also admits a $\cF^Y_t$--adapted representation
    \begin{equation}\label{eq:songsde_gen}
        Y_t=Y_0+\int_0^t \underbrace{-F(Y_s,T-s)+\nabla\log \hat{p}(Y_s,T-s)}_{\text{In \Cref{theo:connection},we call this } F'(Y_s,s)}\dd s  +W_t.
    \end{equation}
\end{theorem}


\Cref{eq:songsde_gen} corresponds to the backward diffusion process from~\citep{song2021a} and, because it is adapted to the filtration $\cF^{Y}$, it represents a valid, and easy to simulate, measurement process.

By now, it is clear how to go from an $\cF^{Y,X}$--adapted filtration to a $\cF^Y$--adapted one. We also showed that a $\cF^Y$--adapted filtration can be linked to the reverse, $\cF^{\hat{Y}}$--adapted process induced by a forward diffusion \gls{SDE}. What remains to be discussed is the connection that exists between the $\cF^Y$--adapted filtration, and its \textit{enlarged} version $\cF^{Y,X}$. In other words, we have shown that a forward, diffusion \gls{SDE} admits a backward process which is compatible with our generative model that simulates a \gls{NLF} process having access only to measurements, but we need to make sure that such process admits a formulation that is compatible the standard \gls{NLF} framework in which latent abstractions are available.

To do this, we can leverage existing results about Markovian bridges~\citep{rogers2000diffusions, ye2022first} (and further work~\citep{aksamit2017enlargement,ouwehand2022enlargement,grigorian2023enlargement,CETIN2016651} on filtration enlargement). This requires assumptions about the existence and well-behavedness of densities $p(y,t)$ of the \gls{SDE} process, defined by the logarithm of the Radon-Nikodym derivative of the instantaneous measure $\PO(Y_t\in \dd y)$ w.r.t. the Lebesgue measure in $\mathbb{R}^N$, $\PO(Y_t\in \dd y)=p(y,t)\dd y$\footnote{Similarly to what discussed in footnote 1, the analysis of the existence of the process adapted to $\cF^Y_t$ is considered in the time interval $[0,T)$ \citep{haussman2}. See also \Cref{sec:technicalnote}.}.
\begin{theorem}\label{theo:connection}
    Suppose that on $\ccU$ the Markov stochastic process $Y_t$ satisfies
    \begin{equation}\label{gendir}
        Y_t=Y_0+\int_0^t F'(Y_s,s)\dd s+W_t,
    \end{equation}
where $\{W_{0\leq t\leq T},\cF^Y_{0\leq t\leq T}\}$ is a Brownian motion and $F$ satisfies the requirements for existence and well definition of the stochastic integral \citep{shreve2004stochastic}. Moreover, let \Cref{assumption1} hold. Then, the same process admits $\ctF_t=\sigma(Y_{0\leq s\leq t},Y_T)$--adapted representation
    \begin{equation}\label{bridge}
        Y_t=Y_0+\int_0^t F'(Y_s,s)+\nabla_{Y_s}\log p(Y_T\g Y_s)\dd s+\beta_t,
    \end{equation}
  where $p(Y_T\g Y_s)$ is the density w.r.t the Lebesgue measure of the probability $\PO(Y_T\g \sigma(Y_s))$, and $\{\beta_{0\leq t\leq T},\ctF_{0\leq t\leq T}\}$ is a Brownian motion.
\end{theorem}

The connection between time reversal of diffusion processes and enlarged filtrations is finalized with the result of \citet{al1987enlarged}, Thm. 3.3, where it is proved how the $\beta_t$ term of \Cref{bridge} is a Brownian motion, using the techniques of time reversals of \glspl{SDE}.

Since $\hat{p}(y,T-t)={p}(y,t)$, the enlarged filtration version of \Cref{eq:songsde_gen} reads 
\begin{equation}\label{anders_them_2}
   Y_t=Y_0+\int_0^t \underbrace{-F(Y_s,T-s)+\nabla_{Y_s}\log p(Y_s\g Y_T)\dd s}_{\text{Equivalent to }H(Y_t,X,t)=-F(Y_s,T-s)+\nabla_{Y_s}\log p(Y_s\g g(X))}  +W_t.
\end{equation}

Note that the dependence of $Y_t$ on the latent abstractions $X$ is implicitly defined by conditioning the score term $\nabla_{Y_s}\log p(Y_s\g Y_T)$ by $Y_T$, which is the ``rendering'' of $X$ into the observable data domain.


Clearly, \Cref{anders_them_2} can be reverted to the starting generative \Cref{eq:songsde_gen} by mimicking the results which allowed us to go from \Cref{eq:param_sde} to \Cref{genh_sde}, by noticing that $\E_\PO[\nabla_{Y_s}\log p(Y_T\g Y_s) \g \cF^Y_t]=0$ (informally, this is obtained since $\int \nabla_{y_s}\log p(y_t\g y_s) p(y_t\g y_s)\dd y_t=\int \nabla_{y_s}p(y_t\g y_s)\dd y_t=0$).

It is also important to notice that we can derive the expression for the mutual information between the measurement process and a sample from the data distribution, as follows
\begin{equation}
\cI(Y_{0\leq s\leq t};\rX)=\cI(Y_{0};\rX)+ \frac{1}{2}\E_{\PO}\left[ \int_0^t\norm{ \nabla_{Y_s}\log p(Y_s)-\nabla_{Y_s}\log p(Y_s\g Y_T)}^2 \dd s \right].
\end{equation}
Mutual information is tightly related to the classical loss function of generative diffusion models.

Furthermore, by casting the result of \Cref{MI_eq} according to the forms of \Cref{eq:songsde_gen,anders_them_2}, we obtain the simple and elegant expression 
\begin{equation}
\cI(Y_{0\leq s\leq t};\rX)=\cI(Y_{0};\rX)+ \frac{1}{2}\E_{\PO}\left[ \int_0^t\norm{ \nabla_{Y_s}\log p(Y_T\g Y_s)}^2 \dd s \right].    
\end{equation}
 
In \Cref{sec:lineardiff}, we present a specialization of our framework for the particular case of linear diffusion models, recovering the expressions for the variance-preserving and variance-exploding \glspl{SDE} that are the foundations of score-based generative models~\citep{song2021a}.

\section{An informal summary of the results}\label{sec:summary}
\begin{figure}
\scalebox{0.8}{
\begin{tikzpicture}[>=stealth, node distance=2cm]

\node(X){\color{blue}$X$};
\node[below= 1cm of X] (Y0) {\color{darkgreen}$Y_0$};
\node[right of= Y0] (Y1) {\color{darkgreen}$Y_1$};
\node[right of= Y1] (Y2) {\color{blue}$Y_2$};
\node[right of= Y2] (Y3) {\color{blue}$Y_3$};

\draw[->] (X) -- (Y0);
\draw[->] (X) -- (Y1);
\draw[->] (X) -- (Y2);
\draw[->,color=blue] (X) -- (Y3);

\node[below =1cm of Y0] (pi0) {\color{darkgreen}$\pi_0$};
\node[below =1cm of Y1] (pi1) {\color{darkgreen}$\pi_1$};
\node[below =1cm of Y2] (pi2) {\color{orange}$\pi_2$};
\node[below =1cm of Y3] (pi3) {$\pi_3$};


\draw[->] (Y0) -- (pi0);
\draw[->,color=darkgreen] (Y0) -- (pi1);
\draw[->,color=darkgreen] (Y1) -- (pi1);
\draw[->] (Y1) -- (pi2);
\draw[->] (Y2) -- (pi2);
\draw[->] (Y2) -- (pi3);
\draw[->] (Y3) -- (pi3);

\draw[->] (Y0) -- (Y1);
\draw[->] (Y1) -- (Y2);
\draw[->,color=blue] (Y2) -- (Y3);

\draw[->,color=darkgreen] (pi0) -- (pi1);
\draw[->] (pi1) -- (pi2);
\draw[->] (pi2) -- (pi3);

\node[below= 1cm of pi0] (piH0)  {$\langle\pi_0,{\gX}\rangle$};
\node[below= 1cm of pi1] (piH1)  {$\langle\pi_1,{\gX}\rangle$};
\node[below= 1cm of pi2] (piH2)  {\color{orange}$\langle\pi_2,{\gX}\rangle$};
\node[below= 1cm of pi3] (piH3) {$\langle\pi_3,{\gX}\rangle$};

\draw[->] (pi0) -- (piH0);
\draw[->] (pi1) -- (piH1);
\draw[->,color=orange] (pi2) -- (piH2);
\draw[->] (pi3) -- (piH3);

\node[draw, rounded corners, color=red, fit=(Y0) (Y3)] (ybox) {};
\node[draw, rounded corners, color=red, fit=(pi3)] (pibox) {};


\node[below= 1.7cm of pi2] (kushner){\color{darkgreen}$\dd \pi_t=\pi_t\left(H-\langle \piphit,H\rangle\right) \left(\dd Y_t-\langle \piphit,H\rangle\dd t\right)$};
\draw[<->, dashed, ,color=darkgreen] (pi1) -- (kushner);

\node[below= .6cm of piH3] (statistic){\color{orange}$\langle\piphit,{\gX}\rangle=\int_{ \cS}{\gX}\dd\piphit$};

\draw[<->,dashed,color=orange] (piH2) -- (statistic);

\node[above= .7cm of Y2] (sde){\color{blue}$\dd Y_t=H(Y_t,X,t)\dd t\\+\dd W_t$};
\draw[<->, dashed, ,color=blue] (Y3) -- (sde);

\node [right= 7cm of X](Xn) {$X$};
\node [below= 1cm of Xn] (Y0n){\color{darkgreen}$Y_0$};
\node [right of= Y0n](Y1n)  {\color{darkgreen}$Y_1$};
\node [right of= Y1n](Y2n)  {\color{blue}$Y_2$};
\node [right of= Y2n](Y3n) {\color{blue}$\rX$};

\draw[->,dashed] (Xn) -- (Y0n);
\draw[->,dashed] (Xn) -- (Y1n);
\draw[->,dashed] (Xn) -- (Y2n);
\draw[->,dashed] (Xn) -- (Y3n);

\node [below =1cm of Y0n](pi0n) {\color{darkgreen}$\pi_0$};
\node [below =1cm of Y1n](pi1n) {\color{darkgreen}$\pi_1$};
\node [below =1cm of Y2n](pi2n) {$\pi_2$};
\node [below =1cm of Y3n](pi3n) {$\pi_3$};

\node[below= 1cm of pi0n] (piH0n)  {$\langle\pi_0,H\rangle$};
\node[below= 1cm of pi1n] (piH1n)  {\color{orange}$\langle\pi_1,H\rangle$};
\node[below= 1cm of pi2n] (piH2n)  {\color{blue}$\langle\pi_2,H\rangle$};
\node[below= 1cm of pi3n] (piH3n) {$\langle\pi_3,H\rangle$};

\draw[->] (Y0n) -- (pi0n);
\draw[->,color=darkgreen] (Y0n) -- (pi1n);
\draw[->,color=darkgreen] (Y1n) -- (pi1n);
\draw[->] (Y1n) -- (pi2n);
\draw[->] (Y2n) -- (pi2n);
\draw[->] (Y2n) -- (pi3n);
\draw[->] (Y3n) -- (pi3n);

\draw[->] (Y0n) -- (Y1n);
\draw[->] (Y1n) -- (Y2n);
\draw[->,color=blue] (Y2n) -- (Y3n);

\draw[->,color=darkgreen] (pi0n) -- (pi1n);
\draw[->] (pi1n) -- (pi2n);
\draw[->] (pi2n) -- (pi3n);


\draw[->] (pi0n) -- (piH0n);
\draw[->,color=orange] (pi1n) -- (piH1n);
\draw[->] (pi2n) -- (piH2n);
\draw[->] (pi3n) -- (piH3n);

\draw[->, bend left] (piH0n) -- (Y1n);
\draw[->, bend left] (piH1n) -- (Y2n);
\draw[->, bend left,color=blue] (piH2n) -- (Y3n);

\node[above= .7cm of Y2n] (sden){\color{blue}$\dd Y_t=\langle \pi_t, H\rangle\dd t+\dd W^{\cF^Y}_t$};
\draw[<->, dashed, ,color=blue] (Y3n) -- (sden);

\node[right =7cm of Xn,rotate=-90] (ineq){$I(Y_{0\leq s\leq t},\gX){\color{red}=\cI(\pi_t;\gX)}{\color{orange}\geq I(\langle\piphit,H\rangle, \gX)}$};

\end{tikzpicture}
}
\caption{Graphical intuition for our results: nonlinear filtering (left) and generative modelling (right).}
\label{fig:summary_gen}
\end{figure}
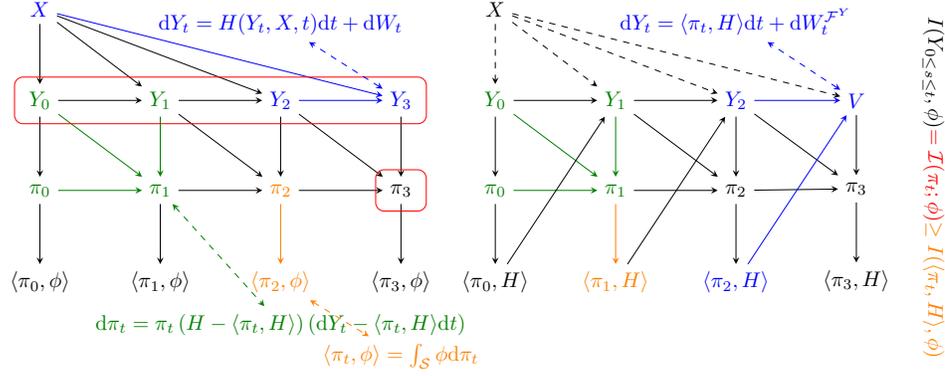

We shall now take a step back from the rigour of this work, and provide an intuitive summary of our results, using \Cref{fig:summary_gen} as a reference.
We begin with an illustration of \gls{NLF}, shown on the left of the figure. We consider an observable latent abstraction $X$ and the measurement process $Y_t$, which for ease of illustration we consider evolving in discrete time, i.e. $Y_0,Y_1,\dots$, and whose joint evolution is described by \Cref{eq:param_sde}. Such interaction is shown in blue: $Y_3$ depends on its immediate past $Y_2$ and the latent abstraction $X$.

The a-posteriori measure process $\pi_t$ is updated in an iterative fashion, by integrating the flux of information. We show this in green: $\pi_1$ is obtained by updating $\pi_0$ with $Y_1-Y_0$ (the equivalent of $\dd Y_t$). This evolution is described by Kushner's equation, which has been derived informally from the result of \Cref{kush}. 
The a-posteriori process is a sufficient statistic for the latent abstraction $X$: for example, $\pi_3$ contains the same information about $\gX$ than the whole $Y_0,\dots,Y_3$ (red boxes). Instead, in general, a projected statistic $\langle\pi_t,{\gX}\rangle$ contains less information than the whole measurement process (this is shown in orange, for time instant 2). The mutual information between all these variables is proven in \Cref{sufficient_statistic}, whereas the actual value of $\cI(Y_{0\leq s\leq t};\gX)$ is shown in \Cref{theo_mi}.

Next, we focus on generative modelling. As by our definition, any stochastic process satisfying \Cref{assumption1} ($Y_3=\rX$, in the figure) can be used for generative purposes. Since the latent abstraction is by definition not available, it is not possible to simulate directly the dynamics using \Cref{eq:param_sde} (dashed lines from $X$ to $Y_t$). Instead, we derive a version of the process adapted to the history of $Y_t$ alone, together with the update of the projection $\langle\pi_t,H\rangle$, which amounts to simulating \Cref{eq:system}.

The update of the upper part of \Cref{eq:system}, which is a particular case of \Cref{kush}, can be \textbf{interpreted} as the composition of two steps: 1) (green) the update of the a-posteriori measure given new available measurements, and, 2) (orange) the projection of the whole $\pi_t$ into the statistic of interest. 
The update of the measurement process, i.e. the lower part of \Cref{eq:system}, is color-coded in blue. This is in stark contrast to the \gls{NLF} case, as the update of e.g. $Y_3=\rX$ does not depend \textbf{directly} on $X$. 
The system in \Cref{eq:system} and its simulation describes the emergence of latent world representations in \gls{SDE}-based generative models:
\begin{tcolorbox}
We interpret the $\cF^Y_t$ measurable quantity $\langle\piphit,H\rangle$ as the cascade of mappings trough the spaces  
\begin{flalign*}
\langle\piphit,H\rangle: & \quad \cC([0,t],\mathbb{R}^N)\rightarrow \cP(\cS)\times \mathbb{R}^N \rightarrow\mathbb{R}^N\\
& \quad Y_{0\leq s\leq t}\rightarrow (\piphit, Y_t) \rightarrow \langle\piphit,H\rangle
\end{flalign*}
We consider it as a mapping that \textbf{first} transforms the whole $Y_{0\leq s\leq t}$ into the \textit{condensed} (in terms of sufficient statistics \Cref{sufficient_statistic}) $\pi_t$, keep also $Y_t$, and \textbf{second} uses these two to construct $\langle\piphit,H\rangle$.
\end{tcolorbox}


The theory developed in this work guarantees that the mutual information between measurements and any statistics $\gX$, grows as described by \Cref{theo_mi}. 
Our framework offers a new perspective, according to which, the dynamics of \gls{SDE}-based generative models~\citep{song2021a} implicitly mimic the two steps procedure described in the box above. We claim that this is the reason why it is possible to dissect the parametric drift of such generative models and find a \textit{representation} of the abstract state distribution $\piphit$, encoded into their activations. Next, we set to root our theoretical findings in experimental evidence.

\section{Empirical Evidence}\label{sec:empirical}
We complement existing empirical studies \citep{park2023understanding,kwon2023diffusionmodelssemanticlatent,chen2023surfacestatisticsscenerepresentations,linhardt2024analysis,tang2023emergent,xiang2023denoising,haas2024discoveringinterpretabledirectionssemantic, sclocchi2024phasetransitiondiffusionmodels} that first measured the interactions between the generative process of diffusion models and latent abstractions, by focusing on a particular dataset that allows for a fine grained assessment of the influence of latent factors. 

\noindent \textbf{Dataset.} 
We use the Shapes3D \citep{kim2018disentangling} dataset, which is a collection of $64 \times 64$ ray-tracing generated images, depicting simple 3D-scenes, with an object (a sphere, cube, ...) placed in a space, described by several attributes (color, size, orientation). Attributes have been derived from the computer program that the ray-tracing software executed to generate the scene: these are transformed into labels associated to each image. In our experiments, such labels are the materialization of the latent abstractions $X$ we consider in this work (see \Cref{appendix:dataset_classes} for details).

\noindent \textbf{Measurement Protocols.} 
For our experiments, we use the base \textsc{ncspp} model described by ~\cite{song2021a}: specifically, our denoising score network corresponds to a \textsc{u-net}~\citep{ronneberger2015}. We train the unconditional version of this model from scratch, using score-matching objective. Detailed hyper-parameters and training settings are provided in~\Cref{appendix:unconditional_training}. Next, we summarize three techniques to measure the emergence of latent abstractions through the lenses of the labels associated to each image in our dataset.
For all such techniques, we use a specific ``measurement'' subset of our dataset, which we partition in 246 training, 150 validation, and 371 test examples. We use a multi-label stratification algorithm \citep{sechidis2011stratification, pmlr-v74-szymański17a} to guarantee a balanced distribution of labels across all dataset splits.
\begin{figure}[htbp]
    \centering
    \includegraphics[width=\linewidth]{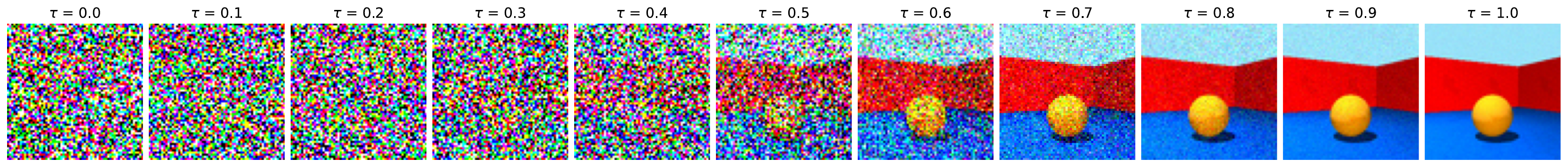}
    \caption{Effect of the variance-exploding schedule on an image corrupted by noise with intensity $\tau$.}
    \label{fig:noise}
\end{figure}

\noindent \textit{Linear probing}. Each image in the measurement subset is perturbed with noise, using a variance-exploding schedule~\citep{song2021a}, with noise levels increasing from 0 to 1 in steps of 0.1, as shown in~\Cref{fig:noise}. 
We extract several feature maps from all the linear and convolutional layers of the denoising score network, for each perturbed image, resulting in a total of 162 feature map sets for each noise level. This process yields 11 different datasets per layer, which we use to train a linear classifier (our probe) for each of these datasets, using the training subset. In these experiments, we use a batch size of 64 and adjust the learning rate based on the noise level (see~\Cref{appendix:linear_probing}). Classifier performance is optimized by selecting models based on their log-probability accuracy observed on the validation subset. The final evaluation of each classifier is conducted on the test subset. Classification accuracy, measured by the model log likelihood, is a proxy of latent abstraction emergence~\citep{chen2023surfacestatisticsscenerepresentations}.

\noindent \textit{Mutual information estimation}. We estimate mutual information between the labels and the outputs of the diffusion model across varying diffusion times, using \Cref{MI_eq_diff} (which is a specialized version of our theory for linear diffusion models, see \Cref{sec:lineardiff}) and adopt the same methodology discussed by \citet{franzese2024minde} to learn conditional and unconditional score functions, and to approximate the mutual information. The training process uses a randomized conditioning scheme: 33\% of training instances are conditioned on all labels, 33\% on a single label, and the remaining 33\% are trained unconditionally. See~\Cref{appendix:mutual_information} for additional details.

\noindent \textit{Forking}. We propose a new technique to measure at which stage of the generative process, image features described by our labels emerge. Given an initial noise sample, we proceed with numerical integration of the backward \gls{SDE}~\citep{song2021a} up to time $\tau$. At this point, we fork $k$ replicas of the backward process, and continue the $k$ generative pathways independently until numerical integration concludes. We use a simple classifier (a pre-trained ResNet50~\citep{resnet} with an additional linear layer trained from scratch) to verify that labels are coherent across the $k$ forks. Coherency is measured using the entropy of the label distribution output by our simple classifier on each latent factor for all the $k$ forks. Intuitively: if we fork the process at time $\tau=0.6$, and the $k$ forks all end up displaying a cube in the image (entropy equals 0), this implies that the object shape is a latent abstraction that has already emerged by time $\tau$. Conversely, lack of coherence implies that such a latent factor has not yet influenced the generative process. Details of the classifier training and sampling procedure are provided in~\Cref{appendix:forking}.
\begin{figure}[htbp]
    \centering
    \includegraphics[width=\linewidth]{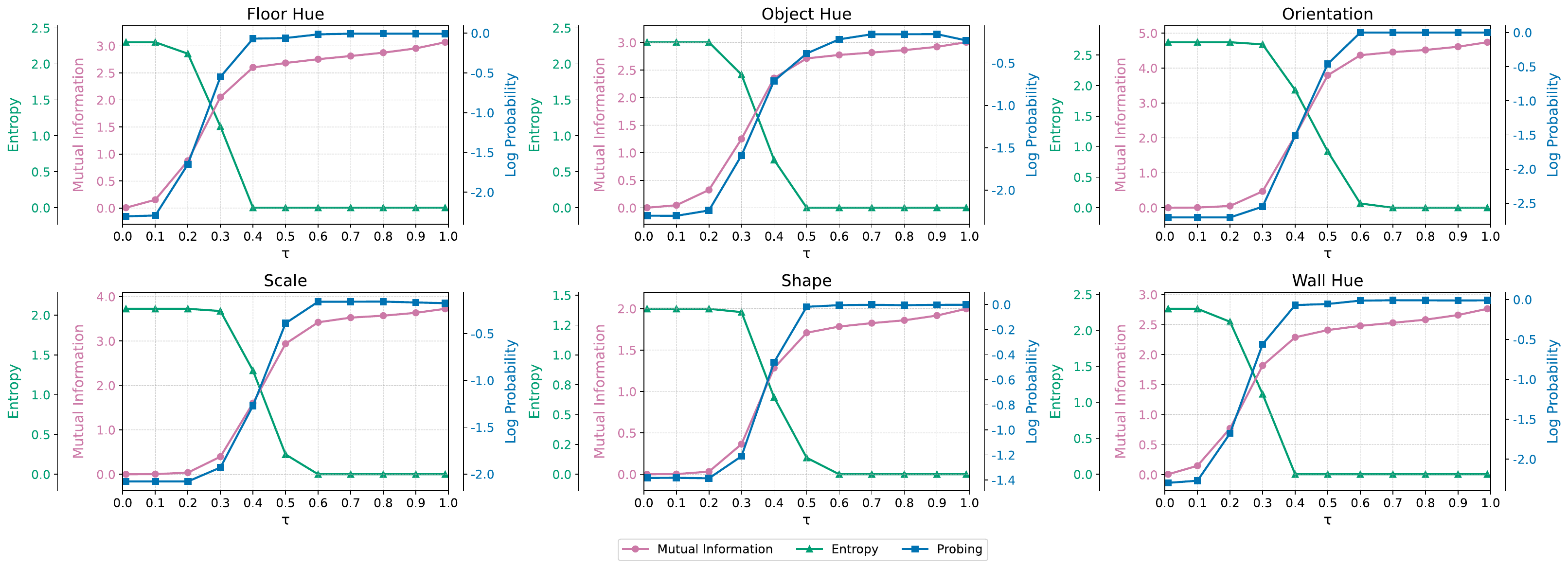}
    \caption{Mutual information, Entropy across forked generative pathways, and Probing results as functions of $\tau$.}
    \label{fig:forking_entropy_mi}
\end{figure}

\noindent \textbf{Results.}
We present our results in \Cref{fig:forking_entropy_mi}. We note that some attributes like \textit{floor hue}, \textit{wall hue} and \textit{shape} emerge earlier than others, which corroborates the hierarchical nature of latent abstractions, a phenomenon that is related to the spatial extent of each attribute in pixel space. This is evident from the results of linear probing, where we evaluate the performance of linear probes trained on features maps extracted from the denoiser network, and from the mutual information measurement strategy and the measured entropy of the predicted labels across forked generative pathways.
Entropy decreases with $\tau$, which marks the moment in which the generative process proceeds along $k$ forks. When generative pathways converge to a unique scene with identical predicted labels (entropy reaches zero), this means that the model has committed to a specific set of latent factors. This coincides with the same noise level corresponding to high accuracy for the linear probe, and high-values of mutual information. Further ablation experiments are presented in \Cref{sec:linearprobing}. 

\section{Conclusion}\label{sec:conclusion}
Despite their tremendous success in many practical applications, a deep understanding of how \gls{SDE}-based generative models operate remained elusive.
A particularly intriguing aspect of several empirical work was to uncover the capacity of generative models to create entirely new data by combining latent factors learned from examples. To the best of our knowledge, there exist no theoretical framework that attempted at describing such phenomenon.

In this work, we closed this gap, and presented a novel theory --- that builds on the framework of \gls{NLF} --- to describe the implicit dynamics allowing 
\gls{SDE}-based generative models to tap into latent abstractions and guide the generative process. Our theory, that required advancing the standard \gls{NLF} formulation, culminates in a new system of joint \glspl{SDE} that fully describe the iterative process of data generation. Furthermore, we derived an information-theoretic measure to study the influence of latent abstractions, which provides a concrete understanding of the joint dynamics.

To root our theory into concrete examples, we collected experimental evidence by means of novel (and established) measurement strategies, that corroborates our understanding of diffusion models. Latent abstractions emerge according to an implicitly learned hierarchy, and can appear early on in the data generation process, much earlier than what is visible in the data domain.
Our theory is especially useful as it allows analyses and measurements of generative pathways, opening up opportunities for a variety of applications, including image editing, and improved conditional generation.
\section{Acknowledgments}
G.F. and P.M were partially funded by project MUSECOM2
- AI-enabled MUltimodal SEmantic COMmunications and COMputing, in the Machine Learning-based Communication Systems, towards Wireless AI (WAI), Call 2022,
ChistERA. M.M. acknowledges the financial support of the European Research Council (ERC) under the European Union’s Horizon 2020 research and innovation programme (AdG ELISA project, Grant agreement No. 101054746).
\clearpage

\bibliography{biblio}
\bibliographystyle{iclr2025_conference}

\newpage

\appendix
\section{Assumptions}\label{app:assumptions}

 \begin{assumption}\label{nullsets}
 Whenever we mention a filtration, we assume as usual that it is augmented with the $\PO-$ null sets, i.e. if the set $N$ is such that $\PO(N)=0$, then all $A\subseteq N$ should be in the filtration.   
\end{assumption}

\begin{assumption}\label{ass0}
    \begin{equation}
        \E_{\PO}[\int_0^t \norm{H(Y_s,X,s)} \dd s]<\infty.
    \end{equation}
\end{assumption}

\begin{assumption}\label{ass0b}
    \begin{equation}
        \PO(\int_0^t \norm{\E_{\PO}[H(Y_s,X,s)\g \cF^Y_s]}^2 \dd s<\infty)=1.
    \end{equation}
\end{assumption}

Eq 2.5 Fundamentals of Stochastic Filtering. Necessary for validity of \Cref{browphi}.
\begin{assumption}\label{ass1}
    \begin{equation}
        \E_{\PO}[\int_0^t \norm{H(Y_s,X,s)}^2 \dd s]<\infty.
    \end{equation}
Note: this assumption implies \Cref{ass0} and \Cref{ass0b}. Despite it is more restrictive, it turns out that it is often easier to check.    
\end{assumption}

Eq 3.19 Fundamentals of Stochastic Filtering. Necessary for validity of \Cref{theo_girsa}.
\begin{assumption}\label{ass2}
    \begin{equation}\label{novikov}
        \E_{\PO}[\exp{\frac{1}{2}\int_0^t \norm{H(Y_s,X,s)}^2 \dd s}]<\infty,
    \end{equation}
    and 
    \begin{equation}\label{cond_novikov}
        \E_{\PO}[\exp{\frac{1}{2}\int_0^t \norm{\E_{\PO}[H(Y_s,X,s)\g  \ctF_s]}^2 \dd s}]<\infty,
    \end{equation}
\end{assumption}
Note: \Cref{ass2}, as well as \Cref{ass1}, are trivially verified when $H$ is bounded.

\section{Proof of \Cref{innovation_theo}}\label{proof_innovation_theo}


We start by combining \Cref{browphi} and \Cref{eq:param_sde}

\begin{flalign*}
     W^{\ctF}_t
     & = Y_0+\int_0^t H(Y_s,X,s) \dd s+ W_t-Y_0-\int_0^t \E_{\PO}(H(Y_s,X,s)\g \ctF_s)\dd s\\& =\int_0^t H(Y_s,X,s) \dd s+ W_t-\int_0^t \E_{\PO}(H(Y_s,X,s)\g \ctF_s)\dd s.
\end{flalign*}

We begin by showing that it is a martingale. For any $0\leq \tau\leq t$ it holds
\begin{flalign*}
    \E_{\PO}[W^{\ctF}_t\g \ctF_\tau] & = \E_{\PO}[\int_0^t H(Y_s,X,s) \dd s\g \ctF_\tau]+\E_{\PO}[W_t\g \ctF_\tau]\\ &\qquad-\E_{\PO}[\int_0^t \E_{\PO}(H(s,Y_s,X)\g \ctF_s) \dd s\g \ctF_\tau]\\&=
    \int_0^t \E_{\PO}[H(Y_s,X,s)\g \ctF_\tau] \dd s+
    \E_{\PO}[\E_{\PO}[W_t\g \cF^{Y,X}_\tau]\g \ctF_\tau]
    \\
    &\qquad -\int_0^\tau \E_{\PO}[H(Y_s,X,s)\g \ctF_s]\dd s-\int_\tau^t \E_{\PO}[H(Y_s,X,s)\g \ctF_\tau]\dd s\\ &=
    \int_0^\tau \E_{\PO}[H(Y_s,X,s)\g \ctF_\tau] \dd s+\E_{\PO}[W_\tau\g \ctF_\tau]+W^{\ctF}_\tau+Y_0-Y_\tau \\& =
\E_{\PO}[\int_0^\tau H(Y_s,X,s)\dd s+W_\tau+Y_0-Y_\tau\g \ctF_\tau]+W^{\ctF}_\tau=W^{\ctF}_\tau.
\end{flalign*}
Moreover, it is easy to check that the cross-variation of $W^{\ctF}_t$ is the same as the one of $W_t$. Then, we can conclude the proof by Levy's characterization of Brownian motion ($W^{\ctF}_0=0$).

\section{Proof of \Cref{theo_girsa}}\label{proof_theo_girsa}

First, by combining the definition of $\psi^{\ctF}_t$ and the fact that $\dd Y_t=\E_{\PO}[H(Y_t,X,t)\g  \ctF_t]+ \dd W^{\ctF}_t$ we obtain

\begin{equation}\label{eq:psi}
     (\psi^{\ctF}_t)^{-1}=\exp(-\int_0^t \E_{\PO}[H(Y_s,X,s)\g  \ctF_s]\dd W^{\ctF}_s-\frac{1}{2}\int_0^t\norm{\E_{\PO}[H(Y_s,X,s)\g \ctF_s]}^2 \dd s ).
\end{equation}
Notice that by \Cref{ass2} (which is actually the usual Novikov's condition), the local martingale $(\psi^\ctF_t)^{-1}$ is a real-valued martingale starting from $(\psi^\ctF_0)^{-1}=1$. Then, we can apply Girsanov theorem and conclude that $\dd \PT = \psi^{\ctF}_T\dd\PO$ is a probability measure under which the process $\{\tilde{W}_{0\leq t\leq T}, \ctF_{0\leq t\leq T}\}$, with
\begin{equation*}
    \tilde{W_t} = W^\ctF_t + \int_0^t \E_{\PO}[H(Y_t,X,s)\g  \ctF_t] \dd s, 
\end{equation*}
is a Brownian motion on the space $\cctU$.

\section{Proof of \Cref{kushner_eq}}\label{proof_kushner_eq}
First, let us give a precise meaning to being a weak solution of \Cref{kush}. We say that $\pi^\ctF_t$ solves \eqref{kush} in a weak sense in, for any for any $\gX:\cS \to \mathbb{R}$ bounded and measurable, it holds
 \begin{equation}\label{weakkush}
    \begin{aligned}
        \scalprod{\pi^\ctF_t}{\phi}& =\scalprod{\pi^\ctF_0}{\phi} \\
        &\quad + \int_0^t \left(\scalprod{\pi^\ctF_s}{H(Y_s,\cdot,s)\phi}-\scalprod{\pi^\ctF_s}{\phi}\langle \pi^\ctF_s,H(Y_s,\cdot,s)\rangle\right) \left(\dd Y_s-\langle \pi^\ctF_s,H(Y_s,\cdot,s)\rangle\dd s\right).
    \end{aligned}
\end{equation}
Let us recall that, on $(\Omega,\cF,\PO)$, the process $Y_t$ has the SDE representation \eqref{eq:param_sde}, where $\{W_{0\leq t\leq T},\cF^{Y,X}_{0\leq t\leq T}\}$ is a Brownian motion. Moreover, by \Cref{theo_girsa} with $\mathcal{R}_t = \cF^{Y,X}_t$, it holds that $\{(Y-Y_0)_{0\leq t\leq T}, \cF^{Y,X}_{0\leq t\leq T}\}$ is a Brownian motion on the space $(\Omega,\cF,\PQ^{\cF^{Y,X}})$, where $\dd \PQ^{\cF^{Y,X}} = (\psi^{\cF^{Y,X}}_T)^{-1} \dd\PO$ and
\begin{equation}\label{def_psi_F}
\psi^{\cF^{Y,X}}_t = \exp(\int_0^t H(Y_s,X,s)\dd Y_s-\frac{1}{2}\int_0^t\norm{H(Y_s,X,s)}^2 \dd s ).
\end{equation}
For notation simplicity, in this subsection $\psi_t^{\cF^{Y,X}}$ and $\PQ^{\cF^{Y,X}}$ are simply indicated as $\pi_t$, $\psi_t$ and $\PQ$ respectively.

Since we aim at showing that \eqref{weakkush} holds, let us fix $\phi$ and let us start from $\E_{\PO}[\gX(X)\g \ctF_t] = \scalprod{\pi^\ctF_t}{\phi}$. 
Bayes Theorem provides us with the following
\begin{equation}\label{kallianpur_striebel}
    \langle\pi^\ctF_t,\gX\rangle=\E_{\PO}[\gX(X)\g \ctF_t]=\frac{\E_{\PQ}[\frac{\dd \PO}{\dd \PQ}\gX(X)\g \ctF_t]}{\E_{\PQ}[\frac{\dd \PO}{\dd \PQ}\g \ctF_t]}=\frac{\E_{\PQ}[\psi_T\gX(X)\g \ctF_t]}{\E_{\PQ}[\psi_T\g \ctF_t]}\defeq \frac{\langle\hat{\pi}^\ctF_t ,\gX\rangle}{\langle\hat{\pi}^\ctF_t ,1\rangle}.
\end{equation}
Starting from the numerator $\langle\hat{\pi}^\ctF_t ,\gX\rangle$, we involve the tower property of conditional expectation and the fact that $\psi_t$ is $\cF^{Y,X}_t$ measurable to write
\begin{flalign}
\scalprod{\hat{\pi}^\ctF_t }{\gX} & = \E_{\PQ}[\psi_T\gX(X)\g \ctF_t]=\E_{\PQ}\left[\E_{\PQ}\left[\psi_T\gX(X)\g \cF^{Y,X}_t\right]\g \ctF_t\right] \nonumber\\
& =  \E_{\PQ}\left[\E_{\PQ}\left[\psi_T\g \cF^{Y,X}_t\right]\gX(X)\g \ctF_t\right]= \E_{\PQ}\left[\psi_t \gX(X)\g \ctF_t\right].\label{bay}
\end{flalign}
Recalling the definition of $\psi_t$ (see \Cref{def_psi_F}), we have
\begin{align}
     \dd \psi_t=\psi_tH(Y_t,X,t) \dd Y_t,
\end{align}
from which it follows
\begin{equation}\label{int_exp}
    \psi_t=1+\int_0^t\psi_s H(Y_s,X,s)\dd Y_s. 
\end{equation}
We continue processing \Cref{bay}, using \Cref{int_exp}, as 
\begin{flalign*}
  \E_{\PQ}\left[\psi_t\phi(X)\g\ctF_s\right] & = \E_{\PQ}\left[ \left(1+\int_0^t\psi_s H(Y_s,X,s)\dd Y_s\right)\gX(X)\g \ctF_t\right]\\
  & = \E_{\PQ}\left[ \gX(X)\g \ctF_t\right]+ \E_{\PQ}\left[\int_0^t \psi_s H(Y_s,X,s)\phi(X)\dd Y_s \g \ctF_t\right]\\
  & = \E_{\PQ}\left[ \gX(X)\g \ctF_t\right]+\int_0^t\E_{\PQ}\left[ \psi_s H(Y_s,X,s)\gX(X)\g \ctF_s\right]\dd Y_s,\\
\end{flalign*}
where to obtain the last equality we used Lemma 5.4 in \cite{xiong}. We also recall that, under $\PQ$, the process $(Y_t - Y_0)$ is independent of $X$. Thus, since $\ctF_t=\sigma(\ctF_0\cup \sigma(Y_{0\leq s\leq t}-Y_0))$ and $ \frac{\dd\PO}{\dd\PQ}\g_{\cF^{Y,X}_0}=1$, we obtain $\E_{\PQ}\left[ \gX(X)\g \ctF_t\right] =\E_{\PO}[ \gX(X)\g \ctF_0]
$. Concluding and rearranging: 
\begin{equation*}
    \scalprod{\hat{\pi}^\ctF_t}{\gX}=\scalprod{\hat{\pi}^\ctF_0}{\gX}+\int_0^t  \scalprod{\hat{\pi}^\ctF_s}{\phi H(Y_s,\cdot,s)}\dd Y_s.
\end{equation*}
Obviously by the same arguments $\scalprod{\hat{\pi}^\ctF_t}{1}=\E_{\PQ}[\frac{\dd \PO}{\dd \PQ}\g \ctF_t]=\E_{\PQ}\left[\psi_t \g \ctF_t\right]$, and 
\begin{equation}\label{eqn: stoc_diff_pi1}
   \scalprod{\hat{\pi}^\ctF_t}{1}=1+\int_0^t  \scalprod{\hat{\pi}^\ctF_s}{ H(Y_s,\cdot,s)}\dd Y_s. 
\end{equation}
From now on, for simplicity we assume that all the processes involved in our computations are $1$-dimensional. The extension to the multidimensional case is trivial. First, let us notice that, by \eqref{eqn: stoc_diff_pi1} and It\^o's lemma, it holds
\begin{equation}
    \dd (\scalprod{\hat{\pi}^\ctF_t}{1}^{-1}) = -\frac{\scalprod{\hat{\pi}^\ctF_t}{ H(Y_t,\cdot,t)}}{\scalprod{\hat{\pi}^\ctF_t}{1}^2}\dd Y_s + \frac{\scalprod{\hat{\pi}^\ctF_t}{ H(Y_t,\cdot,t)}^2}{\scalprod{\hat{\pi}^\ctF_t}{1}^3}\dd t.
\end{equation}
 Then, by the stochastic product rule,
 \begin{align*}
     \dd \scalprod{\pi^\ctF_t}{\psi} & = \dd \left (\scalprod{\hat{\pi}^\ctF_t}{\gX}\scalprod{\hat{\pi}^\ctF_t}{1}^{-1}\right) \\
     & = \scalprod{\hat{\pi}^\ctF_t}{\gX} \dd (\scalprod{\hat{\pi}^\ctF_t}{1}^{-1}) + \scalprod{\hat{\pi}^\ctF_t}{1}^{-1}\dd \scalprod{\hat{\pi}^\ctF_t}{\gX} - \scalprod{\hat{\pi}^\ctF_t}{\phi H(Y_t,\cdot,t)}\frac{\scalprod{\hat{\pi}^\ctF_t}{ H(Y_t,\cdot,t)}}{\scalprod{\hat{\pi}^\ctF_t}{1}^2}\dd t\\
     & = -\scalprod{\hat{\pi}^\ctF_t}{\gX}\frac{\scalprod{\hat{\pi}^\ctF_t}{ H(Y_t,\cdot,t)}}{\scalprod{\hat{\pi}^\ctF_t}{1}^2}\dd Y_t + \scalprod{\hat{\pi}^\ctF_t}{\gX} \frac{\scalprod{\hat{\pi}^\ctF_t}{ H(Y_t,\cdot,t)}^2}{\scalprod{\hat{\pi}^\ctF_t}{1}^3}\dd t\\
     &\quad + \frac{\scalprod{\hat{\pi}^\ctF_t}{\phi H(Y_t,\cdot,t)}}{\scalprod{\hat{\pi}^\ctF_t}{1}}\dd Y_t - \scalprod{\hat{\pi}^\ctF_t}{\phi H(Y_t,\cdot,t)}\frac{\scalprod{\hat{\pi}^\ctF_t}{ H(Y_t,\cdot,t)}}{\scalprod{\hat{\pi}^\ctF_t}{1}^2}\dd t.
 \end{align*}
 Recalling \eqref{kallianpur_striebel} and rearranging the terms lead us to
 \begin{align*}
    \dd \scalprod{\pi^\ctF_t}{\psi} & = -\scalprod{\pi^\ctF_t}{\gX}\scalprod{\pi^\ctF_t}{ H(Y_t,\cdot,t)}\dd Y_t + \scalprod{\pi^\ctF_t}{\gX} \scalprod{\pi^\ctF_t}{ H(Y_t,\cdot,t)}^2\dd t\\
    &\quad + \scalprod{\pi^\ctF_t}{ \phi H(Y_t,\cdot,t)}\dd Y_t - \scalprod{\pi^\ctF_t}{ \phi H(Y_t,\cdot,t)}\scalprod{\pi^\ctF_t}{ H(Y_t,\cdot,t)}\dd t\\
    & = \left(\scalprod{\pi^\ctF_t}{ \phi H(Y_t,\cdot,t)} - \scalprod{\pi^\ctF_t}{\gX} \scalprod{\pi^\ctF_t}{ H(Y_t,\cdot,t)}\right)\left(\dd Y_t - \scalprod{\pi^\ctF_t}{ H(Y_t,\cdot,t)}\dd t\right).
 \end{align*}
\section{Proof of \Cref{theo_mi}}\label{proof_theo_mi}
The proof of this Theorem involves two separate parts. First, we should show the second equality in \Cref{eq:infodef}, i.e. $\int\log\frac{\dd \PO_{\#Y_{0\leq s\leq t},\gX} }{\dd \PO_{\#Y_{0\leq s\leq t}}\dd \PO_{\#\gX}} \dd \PO_{\#Y_{0\leq s\leq t},\gX}= \E_{\PO}\left[ \log\frac{\dd \PO\g_{\ctF_t}}{\dd \PO\g_{\cF^{Y}_t}\dd \PO\g_{\sigma(\gX)}} \right]$. Then, we should prove that the r.h.s of \Cref{eq:infodef} is equal to \Cref{MI_eq}.

\subsection{Part 1}
We overload in this Section the notation adopted in the rest of the paper for sake of simplicity in exposition. A random variable $X$ on a probability space $\ccU$ is defined as a measurable mapping $X:\Omega\rightarrow \Psi$, where the measure space $(\Psi, \cG)$ satisfies the usual assumptions. To be precise, $X$ is measurable w.r.t. $\cF$ if $\forall E\in\cG,X^{-1}(E)\in\cF$, where $X^{-1}(E)=\{\omega\in\Omega:X(\omega)\in E\}$. Equivalently, $\forall E\in\cG,\exists S\in \cF:X(S)=E$.
Of all the possible sigma-algebras which allow measurability, the sigma algebra induced by the random variable, $\sigma(X)$, is the \textit{smallest} one. It can be shown that $\sigma(X) = X^{-1}(\cG) = \{A = X^{-1}(B)\vert B\in\cG\}$. We also denote by $\PO_\# X\colon\cG\to[0,1]$ the push-forward measure associated to $X$ (i.e. the law), which is defined by the relation $\PO_{\# X} (E) = \PO(X^{-1}(E))$ for any $E\in\cG$. Moreover, for any $\cG$-measurable $\phi$, the following integration rule holds
\begin{equation}\label{int_pfwd}
    \int_\Psi \varphi(x)\dd\PO_{\# X}(x) = \int_\Omega \varphi(X(\omega))\dd\PO(\omega).
\end{equation}

Let us focus on $(\Omega,\sigma(X),\PO)$ and let us consider a new measure $\mathrm{Q}$ absolutely continuous w.r.t.~$\PO$. Radon-Nikodym theorem guarantees existence of a $\sigma(X)$-measurable function $Z\colon\Omega\to[0,+\infty)$ (the ``derivative'' $\frac{\dd \mathrm{Q}}{\dd \PO}=Z$) such that $\mathrm{Q}(A)=\int_A Z\dd\PO$, for all $A\in\sigma(X)$. Moreover, by Doob's measurability criterion (see e.g.~Lemma 1.13 in \cite{kallenberg}), there exists a $\cG$-measurable map $f\colon\Psi\to[0,+\infty)$ such that $Z = f(X)$. Then, for any $E\in \cG$,
\begin{flalign*}
    \mathrm{Q}_{\# X}(E)& =\mathrm{Q}(X^{-1}(E))=\int_{X^{-1}(E)}f(X)\dd\PO(\omega)=\int_{\Omega}\mathbf{1}_{X^{-1}(E)}(\omega)f(X(\omega))\dd\PO(\omega)
    \\
    & =\int_{\Omega}\mathbf{1}_E(X(\omega))f( X(\omega))\dd\PO(\omega)=\int_{\Psi}\mathbf{1}_E(x)f(x)\dd\PO_{\# X}(x)= \int_{E}f(x)\dd\PO_{\# X}(x).
\end{flalign*}
In summary, we have that $\frac{\dd\mathrm{Q}_{\#X}}{\dd\PO_{\#X}}=f$, with $f\colon\Psi\to[0,+\infty)$.

Finally, then,
\begin{flalign}
    & \int_\Psi \log(\frac{\dd \PO_{\#X}}{\dd \mathrm{Q}_{\#X}})\dd \PO_{\#X}=-\int_\Psi \log(f)\dd \PO_{\#X}=-\int_\Omega \log(f(X))\dd \PO=\int_\Omega \log\frac{\dd \PO}{\dd \mathrm{Q}}\dd \PO=\E_{\PO}[ \log\frac{\dd \PO}{\dd \mathrm{Q}}].\label{fin_res}
\end{flalign}
What discussed so far, allows to prove that $\int\log\frac{\dd \PO_{\#Y_{0\leq s\leq t},\gX} }{\dd \PO_{\#Y_{0\leq s\leq t}}\dd \PO_{\#\gX}} \dd \PO_{\#Y_{0\leq s\leq t},\gX}= \E_{\PO}\left[ \log\frac{\dd \PO\g_{\ctF_t}}{\dd \PO\g_{\cF^{Y}_t}\dd \PO\g_{\sigma(\gX)}} \right]$. Indeed:
\begin{itemize}
    \item Consider on the space $(\Omega,\ctF_t,\PO\g_{\ctF_t})$ the random variable $T=(Y_{0\leq s\leq t},\gX)$. By construction, $\sigma(T)=\ctF_t$.
    \item Suppose that $\PO\g_{\ctF_t}$ is absolutely continuous w.r.t $\PO\g_{\cF^Y_t}\times \PO\g_{\sigma(\gX)}$ (proved in the next subsection).
    \item Then the desired equality follows from \Cref{fin_res}. 
\end{itemize}

\subsection{Part 2}
Before proceeding, remember that the following holds: for all $\mathcal{R}'_t\subseteq \mathcal{R}_t$, 
$\PT\g_{\mathcal{R}'_t}=\mathrm{Q}^{\mathcal{R}'}\g_{\mathcal{R}'_t}
$. 

We restart from the r.h.s. of  \Cref{eq:infodef}. Thanks to the chain rule for Radon-Nykodim derivatives
\begin{flalign*}
    \log\frac{\dd \PO\g_{\ctF_t}}{\dd \PO\g_{\cF^{Y}_t}\dd \PO\g_{\sigma(\gX)}}& =\log\frac{\dd \PO\g_{\ctF_t}}{\dd \PT\g_{\ctF_t}}\frac{\dd \PT\g_{\ctF_t}}{\dd \PO\g_{\cF^{Y}_t}\dd \PO\g_{\sigma(\gX)}}\\
    & =
   \log\frac{\dd \PO\g_{\ctF_t}}{\dd \PT\g_{\ctF_t}}\frac{\dd \PT\g_{\cF^{Y}_t}}{\dd \PO\g_{\cF^{Y}_t}} \frac{\dd \PT\g_{\ctF_t}}{\dd \PT\g_{\cF^{Y}_t}\dd \PO\g_{\sigma(\gX)}} \\
   & =\log\frac{\dd \PO\g_{\ctF_t}}{\dd \PT\g_{\ctF_t}}\frac{\dd \mathrm{Q}^{\cF^{Y}}\g_{\cF^{Y}_t}}{\dd \PO\g_{\cF^{Y}_t}} \frac{\dd \PT\g_{\ctF_t}}{\dd \PT\g_{\cF^{Y}_t}\dd %
   \PO\g_{\sigma(\gX)}} \\ & =
   \log \psi^{\ctF}_t (\psi^{\cF^{Y}}_t)^{-1} \frac{\dd \PT\g_{\cF^{Y}_t}}{\dd \PT\g_{\cF^{Y}_t}\dd \PO\g_{\sigma(\gX)}} \\ & =
  \log \psi^{\ctF}_t-\log \psi^{\cF^{Y}}_t+\log \frac{\dd \PT\g_{\ctF_t}}{\dd \PT\g_{\cF^{Y}_t}\dd \PT\g_{\sigma(\gX)}},
 \end{flalign*}
where we used \cref{theo_girsa} to make $\psi^\ctF_t$ and $\psi^{\cF^Y}_t$ appear, and the fact that $\dd \PT\g_{\sigma(\gX)} = \dd \PO\g_{\sigma(\gX)}$.

Consequently 
\begin{flalign*}
  & \E_{\PO}\left[\log\frac{\dd \PO\g_{\ctF_t}}{\dd \PO\g_{\cF^Y_t}\dd \PO\g_{\sigma(\gX)}}\right]= \E_{\PO}\left[ \log \psi^{\ctF}_t-\log \psi^{\cF^Y}_t \right]+ \cI(Y_{0};\gX)\\ & =
  \E_{\PO}\left[ \int_0^t \E_{\PO}[h(Y_s,X,s)\g  \ctF_s]\dd W^{\ctF}_s+\frac{1}{2}\int_0^t\norm{\E_{\PO}[h(Y_s,X,s)\g \ctF_s]}^2 \dd s\right]\\& -\E_{\PO}\left[ \int_0^t \E_{\PO}[h(Y_s,X,s)\g  \cF^Y_s]\dd W^{\cF^Y}_s+\frac{1}{2}\int_0^t\norm{\E_{\PO}[h(Y_s,X,s)\g \cF^Y_s]}^2 \dd s\right]+\cI(Y_{0};\gX)\\& = 
  \frac{1}{2}\E_{\PO}\left[ \int_0^t\norm{\E_{\PO}[h(Y_s,X,s)\g \ctF_s]}^2-\norm{\E_{\PO}[h(Y_s,X,s)\g \cF^Y_s]}^2 \dd s\right]+\cI(Y_{0};\gX).
\end{flalign*}

Actually, the result in the main is in a slightly different form. To show equivalence, it is necessary to prove that
\begin{flalign*}
    \E_{\PO}\left[\norm{\E_{\PO}[h(Y_s,X,s)\g \cF^Y_s]}^2\right] & -2\E_{\PO}\left[\E_{\PO}[h(Y_s,X,s)\g \cF^Y_s]\E_{\PO}[h(Y_s,X,s)\g \ctF_s]\right]\\& = -\E_{\PO}\left[\norm{\E_{\PO}[h(Y_s,X,s)\g \cF^Y_s]}^2\right]
\end{flalign*}

which is trivially true since $\E_{\PO}\left[\cdot\g \cF^Y_t\right]=\E_{\PO}\left[\E_{\PO}\left[\cdot\g\ctF_s \right]\g \cF^Y_t\right]$.

\section{Proof of \Cref{sufficient_statistic}}\label{proof:sufficient_statistic}

\subsection{Proof of \Cref{eq:dpi}}\label{proof:eq:dpi}

The inequality is proven considering that: i) \[\cI(Y_{0\leq s\leq t};\gX)=\E_{\PO\g_{\cF^{Y}_t} \times\PO\g_{\sigma(\gX)}}\left[\eta\left(\frac{\dd \PO\g_{\ctF_t}}{\dd \PO\g_{\cF^{Y}_t}\dd \PO\g_{\sigma(\gX)}}\right)\right]\] 
and \[\cI(\tilde{Y}_t;\gX)=\E_{\PO\g_{\sigma(\tilde{Y}_t)}\times\PO\g_{\sigma(\gX)}}\left[\eta\left(\frac{\dd \PO\g_{\sigma(\tilde{Y}_t,\gX)}}{\dd \PO\g_{\sigma(\tilde{Y}_t)}\dd \PO\g_{\sigma(\gX)}}\right)\right]=\E_{\PO\g_{\cF^{Y}_t}\times \PO\g_{\sigma(\gX)}}\left[\eta\left(\frac{\dd \PO\g_{\sigma(\tilde{Y}_t,\gX)}}{\dd \PO\g_{\sigma(\tilde{Y}_t)}\dd \PO\g_{\sigma(\gX)}}\right)\right],\]
with $\eta(x)=x\log x$, ii) that   $\frac{\dd \PO\g_{\sigma(\tilde{Y}_t,\gX)}}{\dd \PO\g_{\sigma(\tilde{Y}_t)}\dd \PO\g_{\sigma(\gX)}}=\E_{\PO\g_{\cF^{Y}_t}\times \PO\g_{\sigma(\gX)}}\left[\frac{\dd \PO\g_{\ctF_t}}{\dd \PO\g_{\cF^{Y}_t}\dd \PO\g_{\sigma(\gX)}}\g \sigma(\tilde{Y}_t,\gX)\right]$ and iii) that Jensen's inequality holds ($\eta$ is convex on its domain)


\begin{flalign*}
&\E_{\PO\g_{\cF^{Y}_t}\times \PO\g_{\sigma(\gX)}}\left[\eta\left(\frac{\dd \PO\g_{\sigma(\tilde{Y}_t,\gX)}}{\dd \PO\g_{\sigma(\tilde{Y}_t)}\dd \PO\g_{\sigma(\gX)}}\right)\right]\\ & \qquad =\E_{\PO\g_{\cF^{Y}_t}\times \PO\g_{\sigma(\gX)}}\left[\eta\left(\E_{\PO\g_{\cF^{Y}_t}\times \PO\g_{\sigma(\gX)}}\left[\frac{\dd \PO\g_{\ctF_t}}{\dd \PO\g_{\cF^{Y}_t}\dd \PO\g_{\sigma(\gX)}}\g \sigma(\tilde{Y}_t,\gX)\right]\right)\right]\\ & \qquad\quad\leq \E_{\PO\g_{\cF^{Y}_t}\times \PO\g_{\sigma(\gX)}}\left[\E_{\PO\g_{\cF^{Y}_t}\times \PO\g_{\sigma(\gX)}}\left[\eta\left(\frac{\dd \PO\g_{\ctF_t}}{\dd \PO\g_{\cF^{Y}_t}\dd \PO\g_{\sigma(\gX)}}\right)\g \sigma(\tilde{Y}_t,\gX)\right]\right]\\ & \qquad =\E_{\PO\g_{\cF^{Y}_t} \times\PO\g_{\sigma(\gX)}}\left[\eta\left(\frac{\dd \PO\g_{\ctF_t}}{\dd \PO\g_{\cF^{Y}_t}\dd \PO\g_{\sigma(\gX)}}\right)\right].\end{flalign*}

\subsection{Proof of Conditional Independence and Mutual Information Equality}
Formally the condition of conditional independence given $\pi$ is satisfied if for any $a_1,a_2$ positive random variables which are respectively $\sigma(X)$ and $\cF^Y_t$ measurable, the following holds: $\E_{\PO}[a_1a_2\g \sigma(\pi_t)]=\E_{\PO}[a_1\g \sigma(\pi_t)]\E_{\PO}[a_2\g \sigma(\pi_t)]$ (see for instance \cite{van1985invariance}).

The sigma-algebra $\sigma(\pi_t)$ is by definition the smallest one that makes $\pi_t$ measurable. Since $\pi_t$ is $\cF^Y_t$ measurable, clearly $\sigma(\pi_t)\subseteq \cF^Y_t$. By the very definition of conditional expectation, $\E_{\PO}[a_1\g \cF^Y_t]=\langle \pi_t,a_1\rangle$, which is an $\sigma(\pi_t)$ measurable quantity. Then $\E_{\PO}[a_1a_2\g \sigma(\pi_t)]=\E_{\PO}[\E_{\PO}[a_1a_2\g\cF^Y_t]\g \sigma(\pi_t)]=\E_{\PO}[\E_{\PO}[a_1\g\cF^Y_t]a_2\g \sigma(\pi_t)]=\E_{\PO}[\E_{\PO}[\langle \pi_t,a_1\rangle a_2\g \sigma(\pi_t)]=\langle \pi_t,a_1\rangle\E_{\PO}[a_2\g \sigma(\pi_t)]$. Since $\langle\pi_t,a_1\rangle=\E_\PO[\langle\pi_t,a_1\rangle\g \sigma(\pi_t)]=\E_\PO[\E_\PO[a_1\g \cF^Y_t]\g \sigma(\pi_t)]=\E_\PO[a_1\g \sigma(\pi_t)]$, the proof of conditional independence is concluded.

In summary, $\sigma(X)$ and $\cF^Y_t$ are conditionally independent given $\sigma(\pi_t)$ ($\subset \cF^Y_t$). This implies that $\PO(A\g \sigma(\pi_t))=\PO(A\g \cF^Y_t),\quad \forall A\in\sigma(X)$, or equivalently $\E_\PO[\mathbf{1}(A)\g \sigma(\pi_t)]=\E_\PO[\mathbf{1}(A)\g \cF^Y_t]$. To prove this, it is sufficient to show that for any $B\in\cF^Y_t$, $\E_\PO[\E_\PO[\mathbf{1}(A)\g \sigma(\pi_t)]\mathbf{1}(B)]=\E_\PO[\mathbf{1}(A)\mathbf{1}(B)]$. By standard properties of conditional expectation $\E_\PO[\E_\PO[\mathbf{1}(A)\g \sigma(\pi_t)]\mathbf{1}(B)]=\E_\PO[\E_\PO[\mathbf{1}(A)\g \sigma(\pi_t)]\E_\PO[\mathbf{1}(B)\g \sigma(\pi_t)]]$. Due to conditional independence $\E_\PO[\mathbf{1}(A)\g \sigma(\pi_t)]\E_\PO[\mathbf{1}(B)\g \sigma(\pi_t)]=\E_\PO[\mathbf{1}(A)\mathbf{1}(B)\g \sigma(\pi_t)]$. Then, $\E_\PO[\E_\PO[\mathbf{1}(A)\g \sigma(\pi_t)]\E_\PO[\mathbf{1}(B)\g \sigma(\pi_t)]]=\E_\PO[\E_\PO[\mathbf{1}(A)\mathbf{1}(B)\g \sigma(\pi_t)]]=\E_\PO[\mathbf{1}(A)\mathbf{1}(B)]$.

The mutual information equality is then proved considering that $\frac{\dd \PO\g_{\ctF_t}}{\dd \PO\g_{\cF^{Y}_t}\dd \PO\g_{\sigma(\gX)}}=\frac{\dd \PO(\omega^x\g{\cF^Y_t})}{\dd  \PO(\omega^x)}$, since the conditional probabilities exist, and that $\PO(\omega^x\g{\cF^Y_t})=\PO(\omega^x\g\sigma(\pi_t))$.
\section{A technical note}\label{sec:technicalnote}
 As anticipated in the main, \Cref{assumption1} might be incompatible with the other technical assumptions in \Cref{app:assumptions}.
 The problem might arise for singularities in the drift term at time $t=T$, which are usually present in the construction of dynamics satisfying \Cref{assumption1} like stochastic bridges. This mathematical subtlety can be more clearly interpreted by noticing that when \Cref{assumption1} is satisfied the evolution of the posterior process $\pi_t$ at time $T$ can occupy a portion of the space of dimensionality lower than at any $T-\epsilon$, $\epsilon>0$. Or, we can notice that if \Cref{assumption1} is satisfied, $ \cI(Y_{0\leq s\leq T};\rX)=\cI(\rX;\rX)$ which can be equal to infinity depending on the actual structure of $\cS$ and the mapping $\rX$. In many cases, a simple technical solution is to consider in the analysis only dynamics of the process in the time interval $[0,T)$\footnote{This is akin to the discussion of \textit{arbitrage} strategies in finance when the initial filtration is augmented with knowledge of the future value at certain time instants, and the fact that while the new process adapted w.r.t the new filtration is also a martingale w.r.t. a given new measure for all $t\in[0,T)$, it fails to do so for $t=T$ (thus giving an arbitrage opportunity).}.In the reduced time interval $[0,T)$, the technical assumptions are generally shown to be satisfied. For the practical purposes explored in this work this restriction makes no difference, and consequently neglect it for the rest of our discussion.

\section{Linear Diffusion Models}\label{sec:lineardiff}

Consider the particular case of \textbf{linear} generative diffusion models \cite{song2021a}, which are widely adopted in the literature and by practitioners. We consider the particular case of \Cref{eq:diffmod_gen}, where the function $F$ has linear expression
\begin{equation}\label{eq:diffmod}
\hat{Y}_t=\hat{Y}_0-\alpha \int\limits_{0}^t \hat{Y}_{s}\dd s+\hat{W}_t,
\end{equation}
for a given $\alpha\geq 0$. We assume of course again that \Cref{assumption1} holds, which implies that we should select $\hat{Y}_0=Y_T=\rX$. Now, $\alpha$ dictates the behavior of the \gls{SDE}, which can be cast to the so called Variance-Preserving and Variance Exploding schedules of diffusion models \cite{song2021a}. In diffusion models jargon, \Cref{eq:diffmod} is typically referred to as a \textit{noising} process. Indeed, by analysing the evolution of \Cref{eq:diffmod}, $\hat{Y}_{t}$ evolves to a noisier and noisier version of $\rX$ as $t$ grows. In particular, it holds that 
$$\hat{Y}_t=\exp(-\alpha t)\rX+\exp(-\alpha t)\int_0^t \exp(\alpha s)\dd \hat{W}_s.$$ 

The next result is a particular case of \Cref{anders_thm}.

\begin{lemma}\label{anders_thm2}
    Consider the stochastic process $Y_t$ which solves \Cref{eq:diffmod}. The same stochastic process also admits a $\cF^Y_t$--adapted representation
    \begin{equation}\label{eq:songsde}
        Y_t=Y_0+\int_0^t \alpha Y_s+2\alpha\frac{\exp(-\alpha (T-s))\E_{\PO}[\rX\g \sigma(Y_s)]-Y_s}{1-\exp(-2\alpha (T-s))} \dd s  +W_t,
    \end{equation}
    where $Y_0=\exp(-\alpha T)\rX+\sqrt{\frac{1-\exp(-2\alpha T)}{2\alpha}}\epsilon$, with $\epsilon$ a standard Gaussian random variable independent of $\rX$ and $W_t$.
\end{lemma}

As discussed in the main paper, we can now show that the same generative dynamics can be obtained under the \gls{NLF} framework we present in this work, without the need to explicitly defining a backward and a forward process. In particular, we can directly select a observation function that corresponds to an Orstein-Uhlenbeck bridge \citep{mazzolo2017constraint,corlay2013properties}, consequently satisfying  \Cref{assumption1}, and obtain the generative dynamics of classical diffusion models. 
In particular we consider the following about $H$\footnote{Notice that with $H$ selected as in \Cref{assumption2} the validity of the theory considered is restricted to the time interval $[0,T)$, see also \Cref{sec:technicalnote}.}:
\begin{assumption}\label{assumption2}
   The function $H$ in \Cref{eq:param_sde} is selected to be of the \textit{linear} form 
   \begin{equation}
       H(Y_t,X,t)=m_t \rX- \frac{\dd \log m_t}{\dd t} Y_t,
   \end{equation}

with $m_t=\frac{\alpha}{\sinh{(\alpha(T-t))}}$, where $\alpha\geq 0$. When $\alpha=0$, $m_t=\frac{\dd \log m_t}{\dd t}=\frac{1}{T-t}$. Furthermore, $Y_0$ is selected as in \Cref{anders_thm}. Under this assumption, $Y_T=\rX,\quad \PO-a.s.$, i.e. \Cref{assumption1} is satisfied \hyperref[proof_eq:analydiff]{[Proof]}.

\end{assumption}

In summary, the particular case of \Cref{eq:param_sde} (which is $\cF^{Y,X}$ adapted) under \Cref{assumption2}, can be transformed into a generative model leveraging \Cref{innovation_theo}, since \Cref{assumption1} holds. When doing so, we obtain that the process $Y_t$ has $\cF^{Y}$ adapted representation equal to 
\begin{equation}\label{gen_sde}
    Y_t=Y_0+\int_0^t m_s \E_{\PO}(\rX\g \cF^{Y}_s)\dd s-\int_0^t \frac{\dd \log m_s}{\dd s} Y_s\dd s+{W}^{\cF^Y}_t,
\end{equation}
which is nothing but \Cref{eq:songsde} after some simple algebraic manipulation. The only relevant detail worth deeper exposition is the clarification about the actual computation of expectation of interest. If $\PO$ is selected such that $\hat{Y}_t$ solves \Cref{eq:diffmod}, we have that 
\begin{equation}
    \E_{\PO}(\rX\g \cF^{Y}_t )=\E_{\PO}( Y_T \g \sigma(Y_{0\leq s\leq t}) )=\E_{\PO}( \hat{Y}_0 \g \sigma(\hat{Y}_{T-t\leq s\leq T}))=\E_{\PO}( \hat{Y}_0 \g \sigma(\hat{Y}_{T-t}))=\E_{\PO}( \rX \g \sigma(Y_t)),
\end{equation}
where the second to last equality is due to the Markov nature of $\hat{Y}_t$.

Moreover, in this particular case we can express the mutual information $\cI(Y_{0\leq s\leq t};\gX)=\cI(Y_{t};\gX)$ ( where we removed the past of $Y$ since the following Markov chain holds $\gX\rightarrow \hat{Y}_0\rightarrow \hat{Y}_{t>0}$) can be expressed in the simpler form
\begin{equation}\label{MI_eq_diff}
      \cI(Y_{t};\gX)= \cI(Y_{0};\gX)+ \frac{1}{2}\E_{\PO}\left[ \int_0^t m^2_s\norm{\E_{\PO}[\rX \g \sigma(Y_s)]-\E_{\PO}[\rX \g \sigma(Y_s, \gX)]}^2 \dd s \right]
\end{equation}
matching the result described in \cite{franzese2023minde}, obtained with the formalism of time reversal of \glspl{SDE}.

\section{Discussion about \Cref{assumption2}}\label{proof_eq:analydiff}
This is easily checked thanks to the following equality \begin{equation}\label{eq:analydiff}
    Y_t=Y_0\frac{m_0}{m_t}+\rX\frac{m_0}{m_{T-t}}+\int_0^t \frac{m_{s}}{m_t}\dd W_s.
\end{equation}

To avoid cluttering the notation, we define $f_t=\ft$. To show that \Cref{eq:analydiff} is true, it is sufficient to observe i) that initial conditions are met and ii) that the time differential of the process is the correct one. We proceed to show that indeed the second condition holds (the first one is trivially observed to be true).

\begin{flalign*}
   & \dd Y_t= -\alpha Y_0\frac{\cosh(\alpha (T-t))}{\sinh(\alpha T)}+\alpha r(X)\frac{\cosh(\alpha t)}{\sinh(\alpha T)}-\alpha\cosh(\alpha (T-t))\int_0^t \frac{1}{\sinh(\alpha (T-s))}\dd W_s+\dd W_t\\ & =
   -\alpha \frac{\cosh(\alpha (T-t))}{\sinh(\alpha (T-t))}\left(Y_0\frac{\sinh(\alpha (T-t))}{\sinh(\alpha T)}+\int_0^t \frac{\sinh(\alpha (T-t))}{\sinh(\alpha (T-s))}\dd W_s\right)+\alpha r(X)\frac{\cosh(\alpha t)}{\sinh(\alpha T)}+\dd W_t\\ & =
   -\alpha \coth(\alpha (T-t))\left(Y_t-r(X)\frac{\sinh(\alpha t)}{\sinh(\alpha T)}\right)+\alpha r(X)\frac{\cosh(\alpha t)}{\sinh(\alpha T)}+\dd W_t\\ & =-f_t Y_t+\alpha r(X)\left(\frac{\coth(\alpha (T-t))\sinh(\alpha t)}{\sinh(\alpha T)}+\frac{\cosh(\alpha t)}{\sinh(\alpha T)}\right)+\dd W_t\\ & =
   -f_t Y_t+\alpha r(X)\left(\frac{\coth(\alpha (T-t))\sinh(\alpha t)+\cosh(\alpha t)}{\sinh(\alpha T)}\right)+\dd W_t\\ & =
   -f_t Y_t+\alpha r(X)\left(\frac{\coth(\alpha (T-t))\sinh(\alpha t)+\cosh(\alpha t)}{\sinh(\alpha T)}\right)+\dd W_t\\ & =-f_t Y_t+m_t r(X)+\dd W_t
\end{flalign*}

where the result is obtained considering that
\begin{flalign*}
    &\frac{\coth(\alpha (T-t))\sinh(\alpha t)+\cosh(\alpha t)}{\sinh(\alpha T)}= \frac{\frac{e^{\alpha (T-t)}+e^{-\alpha (T-t)}}{e^{\alpha (T-t)}-e^{-\alpha (T-t)}}\left(e^{\alpha t}-e^{-\alpha t}\right)+\left(e^{\alpha t}+e^{-\alpha t}\right)}{e^{\alpha T}-e^{-\alpha T}}\\ &\qquad =
    \frac{\frac{e^{\alpha T}+e^{-\alpha (T-2t)}-e^{\alpha(T-2t)}-e^{-\alpha T}}{e^{\alpha (T-t)}-e^{-\alpha (T-t)}}+\left(e^{\alpha t}+e^{-\alpha t}\right)}{e^{\alpha T}-e^{-\alpha T}}\\ & \qquad=
    \frac{e^{\alpha T}+e^{-\alpha (T-2t)}-e^{\alpha(T-2t)}-e^{-\alpha T}+ e^{\alpha T}- e^{-\alpha (T-2t)}+ e^{\alpha (T-2t)}-e^{-\alpha T}}{\left(e^{\alpha (T-t)}-e^{-\alpha (T-t)}\right)\left(e^{\alpha T}-e^{-\alpha T}\right)}\\ & \qquad=\frac{2}{e^{\alpha (T-t)}-e^{-\alpha (T-t)}}.
\end{flalign*}

\section{Experimental Details}
\label{appendix:training_details}

\subsection{Dataset details}
\label{appendix:dataset_classes}

The Shapes3D dataset~\citep{kim2018disentangling} includes the following attributes and the number of classes for each, as shown in Table~\ref{tab:dataset_classes}.

\begin{table}[htbp]
    \centering
    \caption{Attributes and class counts in the Shapes3D dataset.}
    \label{tab:dataset_classes}
    \begin{tabular}{ll}
        \toprule
        \textbf{Attribute} & \textbf{Number of Classes} \\
        \midrule
        Floor hue & 10 \\
        Object hue & 10 \\
        Orientation & 15 \\
        Scale & 8 \\
        Shape & 4 \\
        Wall hue & 10 \\
        \bottomrule
    \end{tabular}
\end{table}

\subsection{Unconditional Diffusion Model Training}
\label{appendix:unconditional_training}

We train the unconditional denoising score network using the NCSN++ architecture \citep{song2021a}, which corresponds to a \textsc{U-Net}~\citep{ronneberger2015}. The model is trained from scratch using the score-matching objective. The training hyperparameters are summarized in Table~\ref{tab:unconditional_training}.

\begin{table}[htbp]
    \centering
    \caption{Hyperparameters for unconditional diffusion model training.}
    \label{tab:unconditional_training}
    \begin{tabular}{ll}
        \toprule
        \textbf{Parameter} & \textbf{Value} \\
        \midrule
        Epochs & 100 \\
        Batch size & 256 \\
        Learning rate & $1 \times 10^{-4}$ \\
        Optimizer & AdamW~\citep{loshchilov2018decoupled} \\
        \quad $\beta_1$ & 0.95 \\
        \quad $\beta_2$ & 0.999 \\
        \quad Weight decay & $1 \times 10^{-6}$ \\
        \quad Epsilon & $1 \times 10^{-8}$ \\
        Learning rate scheduler & Cosine annealing with warmup \\
        \quad Warmup steps & 500 \\
        Gradient clipping & 1.0 \\
        EMA decay & 0.9999 \\
        Mixed precision & FP16 \\
        Scheduler & Variance Exploding~\citep{song2021a}\\
        \quad $\sigma_{\text{min}}$ & 0.01 \\
        \quad $\sigma_{\text{max}}$ & 90 \\
        Loss function & Denoising score matching \citep{song2021a} \\
        \bottomrule
    \end{tabular}
\end{table}

\subsection{Linear Probing Experiment Details}
\label{appendix:linear_probing}

In the linear probing experiments, we train a linear classifier on the feature maps extracted from the denoising score network at various noise levels $\tau$. The training details are provided in Table~\ref{tab:linear_probing}.

\begin{table}[htbp]
    \centering
    \caption{Hyperparameters for linear probing experiments.}
    \label{tab:linear_probing}
    \begin{tabular}{ll}
        \toprule
        \textbf{Parameter} & \textbf{Value} \\
        \midrule
        Batch size & 64 \\
        Loss function & Cross-Entropy Loss \\
        Optimizer & Adam \citep{KingBa15} \\
        Learning rate & \begin{tabular}[l]{@{}l@{}}
            $1 \times 10^{-6}$ for $\tau = 0.9$ or $\tau = 0.99$ \\
            $1 \times 10^{-4}$ for other $\tau$ values \\
        \end{tabular} \\
        Number of epochs & 30 \\
        Inputs & \begin{tabular}[l]{@{}l@{}}
            Feature maps (used as-is in the linear layer) \\
            Noisy images (scaled to $[-1, +1]$) \\
        \end{tabular} \\
        \bottomrule
    \end{tabular}
\end{table}

\subsection{Mutual Information Estimation Experiment Details}
\label{appendix:mutual_information}

For mutual information estimation, we train a conditional diffusion model using the same NCSN++ architecture as before. The conditioning is incorporated by adding a distinct class embedding for each label present in the input image, added to the input embedding along with the timestep embedding. The hyperparameters are the same as those used for the unconditional diffusion model (see Table~\ref{tab:unconditional_training}).

To calculate the mutual information, we use Equation~\ref{MI_eq_diff}, estimating the integral using the midpoint rule with 999 points uniformly spaced in $[0, T]$.

\begin{figure}[htbp]
    \centering
    \includegraphics[width=0.8\linewidth]{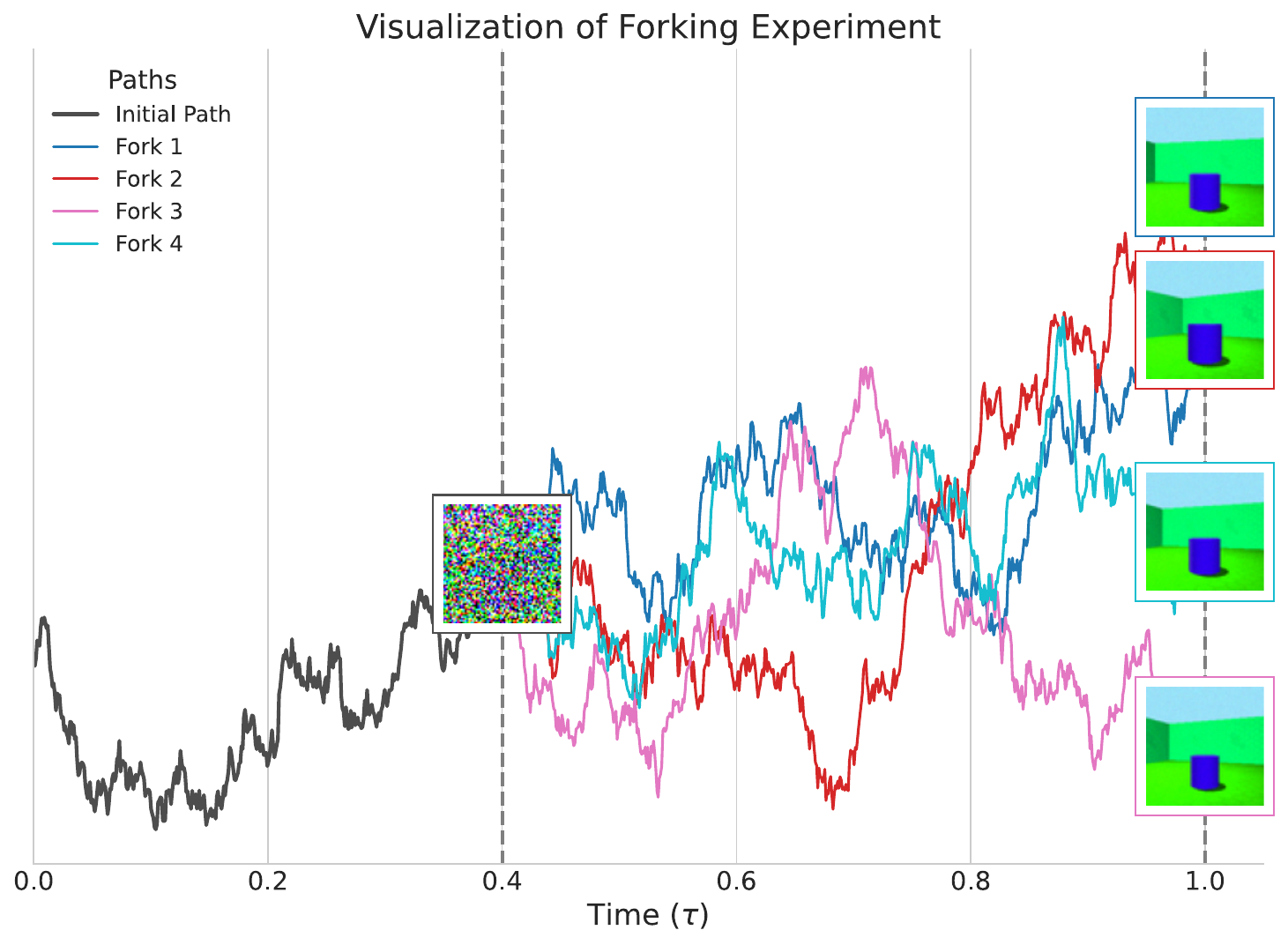}
    \caption{Visualization of the forking experiment with $\texttt{num\_forks} = 4$ and one initial seed. The image at time $\tau = 0.4$ is quite noisy. In the final generations after forking, the images exhibit coherence in the labels \textit{shape}, \textit{wall hue}, \textit{floor hue}, and \textit{object hue}. However, there is variation in \textit{orientation} and \textit{scale}.}
    \label{fig:forking_visualization}
\end{figure}

\subsection{Forking Experiment Details}
\label{appendix:forking}
In the forking experiments, we use a ResNet50 \citep{resnet} model with an additional linear layer, trained from scratch, to classify the generated images and assess label coherence across forks. The training details for the classifier are summarized in Table~\ref{tab:forking_classifier}.

\begin{table}[htbp]
    \centering
    \caption{Hyperparameters for the classifier in forking experiments.}
    \label{tab:forking_classifier}
    \begin{tabular}{ll}
        \toprule
        \textbf{Parameter} & \textbf{Value} \\
        \midrule
        Image size & 224 (resized with bilinear interpolation) \\
        Image scaling & $[-1, +1]$ \\
        Dataset split & \begin{tabular}[l]{@{}l@{}}
            Training set: 72\% \\
            Validation set: 8\% \\
            Test set: 20\% \\
        \end{tabular} \\
        Early stopping & \begin{tabular}[l]{@{}l@{}}
            Stop when validation accuracy exceeds 99\% \\
            Evaluated every 1000 steps \\
        \end{tabular} \\
        Number of epochs & 1 \\
        Optimizer & Adam \citep{KingBa15} \\
        Learning rate & $1 \times 10^{-4}$ \\
        \bottomrule
    \end{tabular}
\end{table}

During the sampling process of the forking experiment, we use the settings summarized in Table~\ref{tab:forking_sampling}.

\begin{table}[htbp]
    \centering
    \caption{Sampling settings for the forking experiments.}
    \label{tab:forking_sampling}
    \begin{tabular}{ll}
        \toprule
        \textbf{Parameter} & \textbf{Value} \\
        \midrule
        Stochastic predictor & Euler-Maruyama method with 1000 steps \\
        Corrector & Langevin dynamics with 1 step \\
        Signal-to-noise ratio (SNR) & 0.06 \\
        Number of forks ($k$) & 100 \\
        Number of seeds & 10 (independent initial noise samples) \\
        \bottomrule
    \end{tabular}
\end{table}

\subsection{Linear probing on raw data}\label{sec:linearprobing}

\begin{figure}[htbp]
  \centering
  \includegraphics[width=\linewidth]{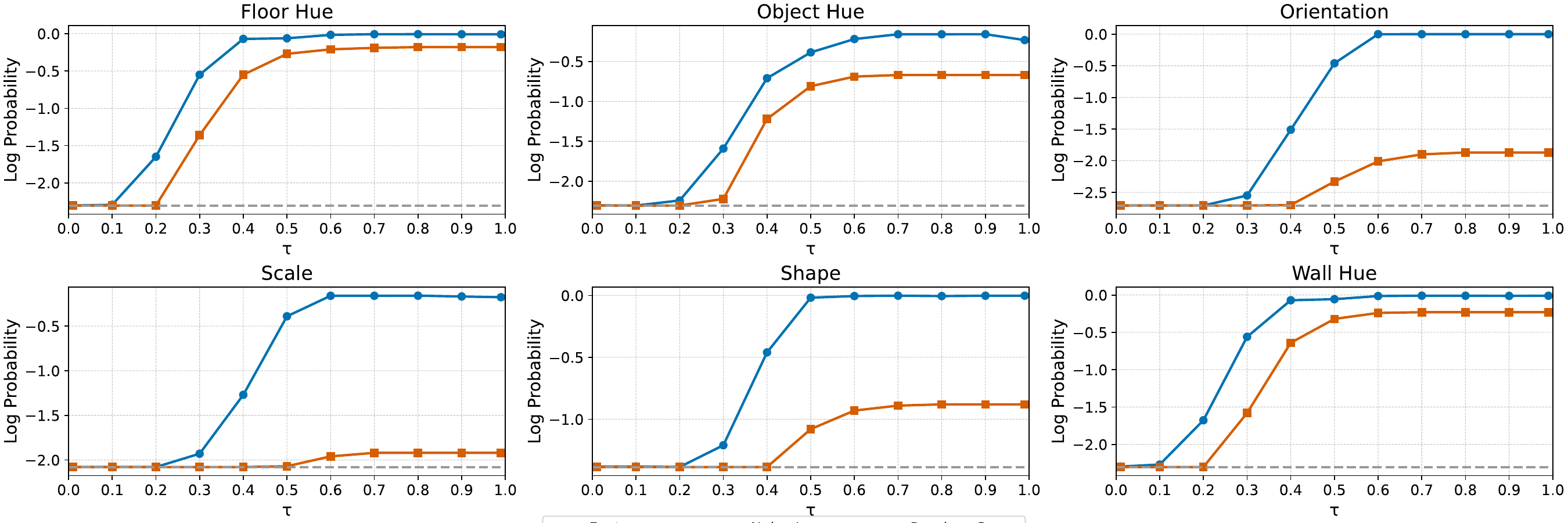}
  \caption{Log-probability accuracy of linear classifiers at $\tau$. 'Feature map' classifiers are trained on network features; 'Noisy Image' trained on noisy images; 'Random Guess' is the baseline for random guessing.}
  \label{fig:probing}
\end{figure}

 In \Cref{fig:probing}, we evaluate the performance of linear probes trained on features maps extracted from the denoiser network, and show compare their log probability accuracy with a linear probe trained on the raw, noisy input and a random guesser. Throughout the generative process, linear probes obtain higher accuracy than the baselines: for large noise levels, a linear probe on raw input data fails, whereas the inner layers of the denoising network extract features that are sufficient to discern latent labels.

\end{document}